# The Anatomy of a Blind Spot: What an LLM Judge Must Be Shown, and How Much

Nitish Kumar† Prateik Pothuneedi† Greg Bandak†
*Hosho* · hoshoai.com
nitish@hoshoai.com · prateik@hoshoai.com · greg@hoshoai.com
†Equal contribution; correspondence to any author. 

**Abstract**

Evaluation of text generated by large language models (LLMs) is increasingly delegated to LLMs acting as judges, yet what such a judge must be shown remains unresolved. We treat a judge whose reports are acted on as a regulator, in the sense of cybernetics, and ask what it must be shown, how much, in what form, and what its reports must get right. On seven public benchmarks we varied only what the judge was shown. Across three judges whose prices differ 10- to 17-fold, what the judge was shown mattered more than which judge it was. When a system revised its answers on feedback from a judge denied the material they had to respect, it ended below its starting point on instruction following and question answering. On instruction following the loss grew with each round, and it persisted when the judge saw the user's request but not the constraints and with a second judge family. On news summarisation, a correct reference summary beside the source undid most of the source's contribution, whereas on single-answer questions a reference helped. About a tenth of the source did almost all of its work, and a small entailment graph matched an LLM judge at finding errors in grounded text. We committed 83 testable clauses before running; of the 81 that ran, 32 met their bar and 50 went in the predicted direction, and most quantitative predictions from the theory failed. We report every shortfall and derive design rules tied to measured effects.

## 1 Introduction

The cost of producing text has fallen sharply, and a growing share of what is written today is written by a language model. This includes finished outputs such as reports, code and answers to questions, and also text that no person ever reads, such as the plans, drafts, tool calls, retrieved passages and rewrites that automated workflows pass from one step to the next. The scale is visible even in published text, where more than half the sentences in a large multi-way parallel web corpus are machine-translated [1]. Within a year of ChatGPT's release, an estimated 17.5% of sentences in computer-science abstracts showed signs of language-model modification [2]. Human review has not kept pace. Each of the corpora we use in this paper required a substantial expert-annotation effort to cover a few thousand outputs, a volume that a deployed system can produce in seconds. As a result, the task of reviewing model output, whether a final answer or an intermediate step, is increasingly given to language models as well.

In a typical deployment, a team checks the output of one language model by asking a second model, the judge, whether that output is good. What the judge sees depends on who built the check. General-purpose judges see the user's prompt and the answer, sometimes with a rubric or a reference answer [3, 4, 5, 6]. Grounding checks for retrieval and summarisation also see the retrieved passages or the source document [7, 8, 9, 10]. Standard judge templates have no place for the rest of the system, such as the system prompt and policy the judged system was meant to follow, or the tools it called. Agent benchmarks mostly score the final outcome rather than the steps that led to it [11, 12, 13]. Which of these inputs a judge actually needs has not been measured [14]. It is reasonable to expect that a missing input limits the judge, and we confirm that it does. We also study the case in which a system acts on the judge's reports, as self-refinement loops, reflective agents and prompt optimisers do, so that the judge is steering the system [15, 16, 17, 18].

We find that this steering can do harm. On two of the three corpora where we closed the loop, a system that revised its output in response to a judge denied the material that output had to respect ended up below where it started, and on the third the loss was within noise. On instruction following the system also ended up below a copy of itself that received no feedback at all, and each further round of feedback pushed it lower. It is already known that language models do not reliably correct their own reasoning without dependable external feedback [19, 20, 21]. Our result adds that feedback from a judge that cannot see that material can be worse than none, and that the damage accumulates. On instruction following and arithmetic, the judge we call blind was not shown the task itself, a stricter condition than a deployed judge that sees the user's prompt but not the system prompt or policy. On question answering it saw the question and lacked only the source knowledge, which is the deployed case, and the system still ended below its start (§8). In an exploratory arm the judge saw the user's request but not the constraints, as if they sat in a system prompt, and did nearly as much damage, and the harm replicated with a second judge family (§8.2). The same pattern appears in training when a model is optimised against an imperfect reward model [22, 23, 24].

Three further findings bear on what a judge should be shown. The first concerns the reference answer. Many pipelines already hold a correct answer for each question [3, 6], so it is tempting to give the judge that answer in place of the source. On news summarisation, where many different summaries are acceptable, we find that this backfires, and that a reference placed beside the source undoes most of what the source had gained. The second, concerning cost, is that a judge priced at 10 to 17 times the cheapest one we tested does not close the gap left by the missing source. The third, concerning quantity, is that the judge does not need the whole source, because about a tenth of it, the part nearest the claims being checked, does almost all the work.

Together these results turn the design of a judge into four questions that measurement can answer: which material must reach the judge, how much of it, in what form, and what its reports must get right once a system acts on them. We answer each of them in this paper, and the answers depend on how the judge is connected to the task more than on which judge is used or how it is tuned.

A single example shows what the judge is missing. It comes from RAGTruth [25], a corpus in which human annotators marked the errors in answers written by retrieval-augmented assistants, systems that first retrieve passages relevant to a question and then answer from them. A user asked for the average cost of adding a second floor to a house. The system retrieved three passages and answered that, based on the provided passages, the cost could not be determined. The annotators marked this answer as contradicting its source. We showed a judge the answer alone and asked, in three independent runs, whether it contained an error, and each time it found nothing wrong. The answer is fluent and careful, and it declines to guess, which is what a cautious system should do when information is missing. When the same judge was also shown the three passages, it flagged the answer. In its words, passage 1 "provides a range for the cost of a second floor addition ($100-$300 per square foot, potentially up to $400-$500 per square foot)". The error cannot be seen in the answer by itself and becomes visible only when the answer is compared with a passage the judge was not shown in the first condition.

This limit has a precise description in the theory of regulation set out by W. Ross Ashby [26]. Ashby was one of the founders of cybernetics, the study of how machines and organisms keep themselves under control, and he classified regulators by where their information comes from. A regulator that senses only the error, the deviation after it has happened, he called error-controlled, and he showed that its success "can at best be partial" ([26], S.12/5). In later work with Roger Conant he set it against the cause-controlled regulator, which draws its information directly from the cause of the disturbance, and for which alone "the regulation may, in principle, be made perfect" ([27], §3).

A judge shown only the output is an error-controlled regulator by construction. It can notice that an answer reads oddly, but it cannot notice that the answer contradicts a passage it has never seen. No capability inside the judge can recover, from its input alone, a distinction that the input does not carry, a limit that information theory states as the data-processing inequality [28]. The judge's own world knowledge is a second channel, but a partial and unreliable one, as the capability ladder of §4 shows. We use the term world, written W, for everything the output was supposed to respect: the retrieved passages, the source article, the instruction, the knowledge an agent was given, and in general the policy and the steps that produced the output (Figure 1). Every experiment in this paper changes one connection, namely whether the judge is wired to the world. A blind judge is shown the output, and on some corpora the question, whereas an informed judge is shown the same material and the world. Where the instruction or the problem statement is itself the world, as on IFEval and GSM8K, the blind judge is not shown the task at all (§3.3). The model, the prompt and the scoring are the same in both. Under RAGTruth's own scoring protocol, which needs no matching step of ours, the most capable judge shown only the answer scores below a mid-tier judge shown the source (F1 0.217 against 0.336). On LLM-AggreFact, where the corpus labels score the judge directly, the source raises balanced accuracy, the mean of the share of defective items the judge flags and the share of error-free items it leaves alone (0.5 is chance), from 0.565–0.587 to 0.775–0.795 across the same three judges.

**Key terms.**

| Term | Meaning |
|---|---|
| Judge | An LLM, or any program, that reads a system's output and reports whether it is acceptable and what is wrong with it. A judge whose reports are acted on is a regulator. We write evaluator where a point holds for any such instrument, language model or not. |
| The world W | Everything the output was supposed to respect: the retrieved passages, the source article or source sentence, the instruction, or the knowledge given to the agent. Its textual core, the material the output was generated from, is the grounding context of the paired experiments. |
| Blind and informed judge | The same judge shown the output without the world (blind), or the output together with the world (informed). Exactly what accompanies the output in the blind arm varies by corpus and is listed in §3.3, but it never includes the world. This is the single variable manipulated in the primary design. |
| Regulator | Ashby's term for any mechanism that keeps a system's output within acceptable bounds [26, 27]. An error-controlled regulator senses only the outcome; a cause-controlled regulator also senses the disturbance that produced it. |
| Closed loop | A design in which the judge's report is returned to the system, the system revises its own output, and a program checks the revision. |
| Marked segment | A span of output a human annotator marked as defective. It is the unit of the paired tests, and no claim here rests on us deciding what counts as a defect. |
| Matcher | The component deciding whether a claimed defect in the judge's report points at a marked segment. There are two routes: a strict one based on quote overlap, and a generous one in which a small model decides. Because the two disagree, we report both (§3.6), and in §4 we lead with results that need no matcher. |

| Term | Meaning |
|---|---|
| Recall and precision | Recall is the share of real defects the judge finds. Precision is the share of its complaints that point at a real defect. Precision needs a class balance that some corpora do not fix, so figures on those corpora plot a false-positive count per error-free answer, or the share of error-free text left unflagged (specificity), and the axis says so. |
| Prediction | A claim with a numerical pass or fail bar, written to a file and committed by hash in our private repository before the run it governs. These are internal pre-commitments, not a public pre-registration (§3.8). We wrote 37, with 83 committed clauses; Table 6 gives the outcome of each. |
| Balanced accuracy | The mean of the share of defective items the judge flags and the share of error-free items it leaves alone. A score of 0.5 is chance. We use the metric wherever a corpus supplies a label per sentence or claim rather than a marked span. |
| Structural judge | A non-generative instrument built from an entailment model. It receives the world but contains no language model, and the graph it builds over output and world is the entailment graph of the figures (§3.4). |
| Certificate | A defect report that names the specific evidence behind each claim. The structural judge emits one for every defect it finds (§3.4). In the closed-loop experiments the checker itself emits a perfect certificate, the ground-truth list of which requirements actually failed (§3.5). |

We organise the paper around six research questions, which are listed with their answers in Table 1. The first establishes that what the judge is shown limits it more than how good the judge is. The next three determine what it is shown by asking which material must reach the evaluator, how much of it and in what form, and whether the evaluator must be a language model at all. The fifth asks what the evaluator's reports must satisfy once a system acts on them and the judge is working as a regulator, and it yields the sharpest result of the paper (§8). The last asks whether the same conditions govern the system being judged. Together the six give a sufficiency standard, under which an evaluator is adequate for a task when what it is shown lets it tell apart the cases that deserve different verdicts. We measure how far below the full world this sufficiency point lies.

**Table 1:** The six questions that organise the paper, each with its answer and the section that reports the measurements behind it.

| Question | What we found | Where |
|---|---|---|
| RQ1. Is what the judge is shown, or the judge itself, the binding constraint? How much does an output-only judge miss, and does more capability recover it? | Capability does not recover it. Under RAGTruth's own protocol, the most capable judge without the source scored below a mid-tier judge with it (F1 0.217 against 0.336). On LLM-AggreFact, scored from the corpus labels, the best blind judge reached a balanced accuracy of 0.587 and the worst informed judge 0.775. | §4 |

| Question | What we found | Where |
|---|---|---|
| RQ2. Which material must the judge be given: anything that looks like useful context, or only what the output was generated from? | Only the output's own source material. The most helpful-looking substitute did the most harm. On news summarisation, a correct reference summary placed beside the true article removed 82% of the precision the article had bought. On its own, the reference was no better than the blind judge (balanced accuracy 0.552 against 0.565). A plausible but wrong document of the same length raised false positives from 12.4 to 20.9 per error-free answer. | §5 |
| RQ3. How much of that material does the judge need, and in what form must it arrive? | How it arrives matters more than how much of it there is. Burying the true source among ten irrelevant documents cost 2 to 5 points of recall. The most relevant 11% of the source kept 98% of the recall of the full source. Numbering the source sentences and requiring a citation for each complaint cut false positives from 3.36 to 2.88. A factorial over four inputs separated their effects. For the language-model judge, the source accounted for the whole gain in detection (+0.322 recall), although the committed four-factor prediction failed. Stating the requirements produced the largest fall in false positives (−3.59 per error-free answer). Showing the question improved neither. | §5, §6 |
| RQ4. Does the evaluator have to be a language model? | Not for grounded text. Given the source and organised as a claim-to-evidence graph, a small entailment model, which only checks whether a statement follows from a text, matched the informed language model at finding errors (recall 0.927 against 0.915) with a quarter fewer false positives. It fails where the output paraphrases its source, as in summarisation (balanced accuracy 0.522). | §7 |
| RQ5. When nobody reads the judge's report and the system instead acts on it by rewriting its answer, what decides whether the answer gets better or worse? | Whether the complaints are correct, not how many are found. Acting on the reports of a judge denied the material the answer had to respect left the system below where it started on instruction following and question answering, while on arithmetic the loss was within noise. On instruction following it did worse than sending no report at all. Constraints passing fell from 84% to 70%, whereas telling the model only to revise reached 87%, and the loss compounds over rounds. The most precise regulator, a ground-truth defect list, reached 90%. | §8 |
| RQ6. When an agent fails, was it the agent or the material it was handed? Do the same rules govern the generator? | They do. On HaluEval QA, two agents (Llama-3.1-8B and Claude Haiku 4.5) answered with accuracy 0.316 and 0.446 given no material, 0.780 and 0.812 given half of it, and 0.926 and 0.948 given all of it. Given irrelevant material they scored 0.200 and 0.204, below their scores with none. | §9 |

This paper makes four contributions.

(1) A closed-loop measurement against a control that sends no report, showing that reports from a judge denied the world leave the system below its starting point on two of three corpora, with the third within noise. On one corpus they leave it below that control, and the loss compounds over rounds.

(2) A controlled measurement of an evaluator's information as a causal variable, on seven public datasets, with 37 predictions, comprising 83 committed clauses, written before their runs and every shortfall reported. It shows that what a judge is shown binds harder than which judge it is.

(3) A sufficiency framework for evaluators that reads Ashby's requirements for a regulator as the four design questions above and states which of those requirements can be measured in an output-only judge and which cannot.

(4) Design rules with measured effect sizes, and the finding that the material a judge actually needs is far narrower than the full world.

The answers themselves are in Table 1, and the design rules they support are in Table 5, each with its effect size. The predictions were committed experiment by experiment. Most of the quantitative predictions derived from the theory failed (§10.2), so the design rules rest on the measurements rather than on the framing.

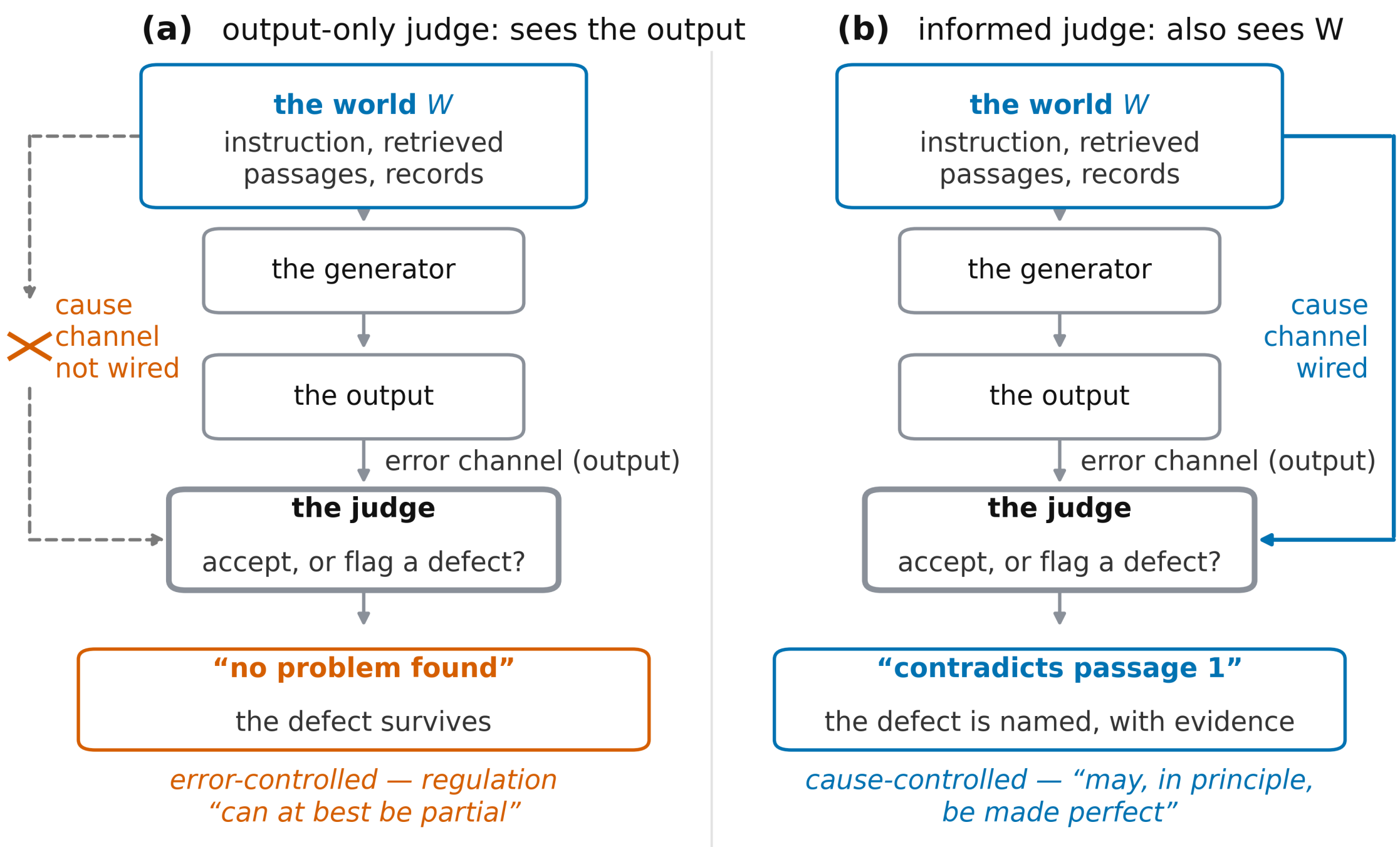

**Figure 1:** An output-only LLM judge is Ashby's error-controlled regulator. Two wirings of the same judge. (a) Output-only wiring: only the output reaches the judge, an error-controlled regulator. (b) This paper: the world W is wired in as well. The panels differ in exactly one connection, the variable every experiment manipulates.

# 2 What a regulator must possess

## 2.1 Ashby's requirements

W. Ross Ashby set out a theory of regulation in An Introduction to Cybernetics [26], and with Roger Conant proved its central theorem fourteen years later [27]. The theory applies to any mechanism that acts to hold some quantity within bounds, such as a thermostat or a governor, and we argue here that an evaluator is another such mechanism. In Ashby's notation a regulator R acts against a disturbance D. The disturbance pushes a system's essential variables, the quantities that must stay in range, out of a goal set, the states that count as acceptable, and the regulator acts to keep them inside it.

Three of Ashby's results bear on evaluation. First, a regulator that senses only the outcome and not the disturbance is error-controlled, and Ashby proves that such regulation can at best be partial. A regulator that senses the disturbance is cause-controlled and may in principle be made perfect. Second, the law of requisite variety holds that a regulator can absorb only as much disturbance as it has distinct states to respond with, so that variety in the regulator is what destroys variety in the outcome. Third, the good regulator theorem holds that an optimal regulator must contain a model of the system it regulates. In Conant and Ashby's construction that model is a partition that separates any two situations calling for different responses. Read as claims about evaluation, these results say that a judge shown only the output is error-controlled by construction and that what it can discriminate is bounded by what its input carries. Such a judge must also be able to tell apart situations that deserve different verdicts.

An evaluator acts as the sensing half of a regulator, separating acceptable output from failed output so that the failures can be corrected. Having identified the output-only judge in §1 as an error-controlled regulator, we set out in this section what a sufficient one must possess. Ashby states his requirements precisely enough that each can be checked against an output-only judge. They fall into three groups: the essential variables and the goal set must be given from outside, information about the disturbance together with variety and capacity must arrive through channels, and a model of the system must exist inside the judge. The middle group is the one a designer controls at deployment time, and it is where we concentrate our measurements. Table 2 lists the requirements in Ashby's words, together with the status of each in a typical language-model judge and the section in which we test it.

**Table 2:** Ashby's requirements for a regulator, the status of each in a typical language-model judge, and where this paper tests it. Quotations are verbatim from An Introduction to Cybernetics [26] except where marked [27]; references of the form S.12/5 are Ashby's own section numbers.

| Requirement | Ashby, verbatim | Status in a typical LLM judge | Where tested here |
|---|---|---|---|
| 1. The essential variables and the goal set, stated first | "The essential variables E are given, and also given is the set of states η in which they must be maintained … These two must be given before all else. Before any regulation can be undertaken or even discussed, we must know what is important and what is wanted." (S.12/1) | a generic quality rubric; no per-task statement of what would count as acceptable [3, 29] | §8: stating the acceptability criterion cuts the judge's false-positive rate on satisfied constraints from 0.553 to 0.405 and narrows what it acts on, without improving the net outcome (P37) |
| 2. Information about the disturbance, not only about the error | "A fundamental property of the error-controlled regulator is that it cannot be perfect … the more successful R is in keeping E constant, the more does R block the channel by which it is receiving its necessary information." (S.12/5). In Conant and Ashby's formulation, when the regulator draws its information directly from the disturbance "the regulation may, in principle, be made perfect" [27] | missing. Output-only judging is error-controlled regulation by construction | §4–§5: the whole blind-versus-informed contrast, and §5's finding that a gold answer, pure error control, is no better than no world |
| 3. Requisite variety | "only variety in R can force down the variety due to D; variety can destroy variety." (S.11/7) | unstated and unmeasured | §8: showing the judge none, 63% or all of an instruction's constraints moves the loop from −0.120 to +0.042 to +0.048 |

| Requirement | Ashby, verbatim | Status in a typical LLM judge | Where tested here |
|---|---|---|---|
| 4. Capacity in the channel the reports travel through | "The law of Requisite Variety says that R's capacity as a regulator cannot exceed R's capacity as a channel of communication." (S.11/11; Ashby's own correspondence with Shannon's Theorem 10 [30]) | often a single scalar score [3], at most $\log_2 5 \approx 2.32$ bits per response on a 1–5 scale; the scores emitted in §8.2 carry 1.56 | §8: dialling the channel from a scalar score to free text to typed ids to a certificate orders the arms by report accuracy, not by bits. The capacity bound is necessary but far from sufficient (P36) |
| 5. A model of the system: the mapping h | "The simplest optimal regulator R of a reguland S produces events R which are related to the events S by a mapping $h : S \rightarrow R$." [27] | absent | §5: a placebo world, and §9: an irrelevant or false world for the agent, which is variety without a model and produces unsupported flagging rather than detection |

Requirement 5 can be stated formally. The theorem of Conant and Ashby makes a more specific statement than its title suggests. The simplest optimal regulator is a deterministic, many-to-one map from system states to regulator states. Two system states may share a regulator state if and only if they need not be told apart for the outcome, so the model only needs to partition the world. To transfer this from acting to judging, let w be the world of a task, y an output, and a(y, w) whether that output is acceptable. A map h is adequate when $h(w) = h(w')$ implies $a(y, w) = a(y, w')$ for every y. Two worlds that share a verdict can be merged at no cost, so an adequate model may still compress the world W heavily. Figure 2 illustrates the condition using the example from §1. Because the verdict differs between the world in which the passage is silent on cost and the world in which it gives a price range per square foot, an adequate model must separate them. A map that merges these two worlds has to issue one verdict for both and so is wrong in one of them. By the data-processing inequality [28], no evaluator downstream of that map can recover the difference. The blind judge implements a map of this kind, so on this account the deficit we measure in §4 should arise in the input partition, before any capability inside the judge is applied.

We note two limits on this argument. The theorem minimises the entropy of the outcome rather than its badness, and we do not estimate any information-theoretic quantity here. In particular, the blind balanced accuracies of the three judges in §4 lie within 0.022 of one another, but that closeness shows a plateau over a narrow range and does not establish a bound on what a blind judge can achieve.

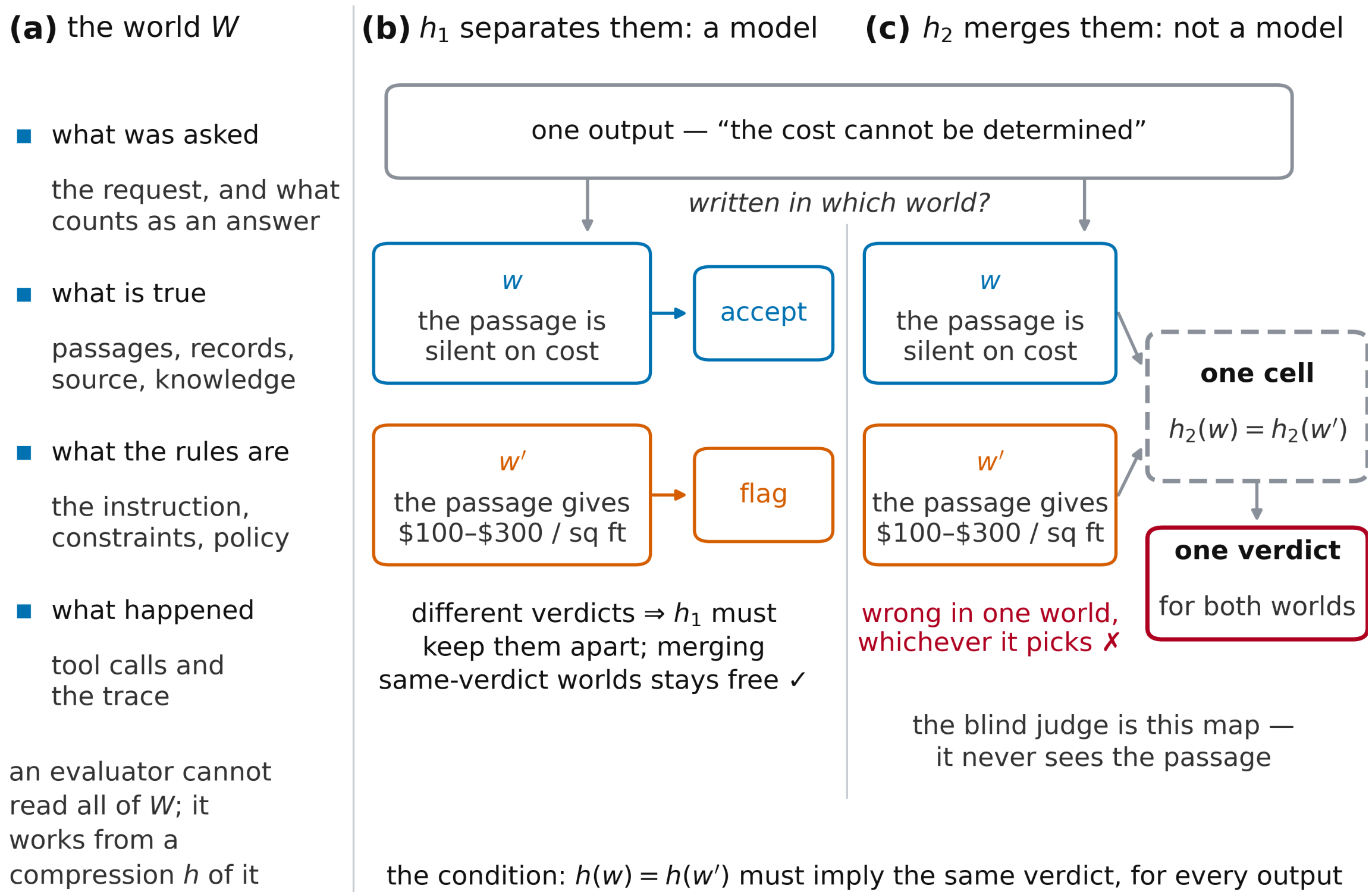

**Figure 2:** A model must separate worlds that get different verdicts. The panels illustrate the model condition with one case. (a) One output and two worlds it could have been written in: w, the passage is silent on cost; w′, it gives a price range per square foot. (b) A model separates them and issues each its verdict. (c) A map that merges them puts both in one cell, with one verdict for both.

## 2.2 Related work

Recent work on world models has since proved a stronger and independent result from the agent's side, namely that agents that generalise across multi-step goals must contain approximate models of their environment and that those models are recoverable from their policies [31]. The authors of that result reject the interpretation that Conant and Ashby placed on their own theorem, holding that it shows only that an entropy-minimising agent has a deterministic policy. For this reason we cite [31] as an independent result and not as a restatement of [27]. The theorem has also been recast as a world-model framework for intelligent systems [32], and a later paper builds on [31] to give it an embodied treatment in the form of a Bayesian analogue [33].

Prior work on LLM evaluation can be organised along the two axes this paper uses, namely what does the evaluating and what the evaluator is allowed to see, which in Ashby's terms are the regulator itself and its input. Each line of work varies one of these axes at a time. Studies of judge bias [3, 34] vary the judge and examine position, verbosity and self-preference. Zheng et al. [3] also report reference-guided grading, in which the judge is handed a reference solution. Large-scale reliability and stability studies [35, 36, 37] measure agreement, consistency and robustness under perturbation and pressure. None of these makes the evidence shown to the judge the manipulated variable.

The second axis, what the evaluator is shown, is closer to our design. ContextualJudgeBench [14] provides the judge with evidence as part of the task definition, so context is not the manipulated variable and the benchmark cannot measure its value. Agent-as-a-Judge [11] extends the view to the execution trace, but the wider world arrives there as part of a new system and is not ablated in a paired comparison of the same judge. Source-conditioned metrics [29, 38] grade the output against its source, while reference-guided grading [3, 6] gives the evaluator a reference answer and treats it as evidence. In §5 we find that the two are not interchangeable, since a gold reference behaves like a wrong world and not like a substitute source. NLI-based consistency checkers [7, 8, 39] study one configuration, output plus source with an entailment model, without comparison across world states or instruments. Khan et al. [40] and Kenton et al. [41] compare judges that can and cannot read the underlying article, in debate and consultancy protocols on a single reading-comprehension task. A recently proposed evidentiary-adequacy criterion for scalable oversight [42] states the standard but does not supply measurements, and our study is an attempt to meet it. To our knowledge, none of these designs varies the evaluator's information as the sole manipulated variable across several corpora and instruments while holding the evaluator fixed. The with- and without-article comparisons of Khan et al. [40] and Kenton et al. [41] come closest, on one task. We describe such a design in §3.

Four further lines of related work are close to our design. The self-correction literature returns a critique to the generating model and measures the revision. It includes Self-Refine [15], Reflexion [16], and the negative result on self-correction without an external signal [19], which we extend in §8. This literature varies the source of the critique rather than the information available to the critic, and it rarely includes a no-feedback control of the kind we include in §8. Retrieval-augmented evaluation frameworks such as RAGAS [9] and ARES [10] already show the judge the retrieved context, so the output-only wiring of §1 is not universal. In these frameworks the context is fixed and not ablated, so they cannot measure what it contributes or when it harms. The TRUE benchmark [43], which consolidated NLI-based factual consistency across eleven datasets, is the direct predecessor of the entailment instrument in §7 and of LLM-AggreFact. Reward-model over-optimisation [22], noted in §1, is the same problem on the training side, where a proxy evaluator scores outputs without access to the world it was meant to track. When training optimises against that evaluator, the quantity the evaluator scores degrades, which corresponds to the compounding loss we observe in §8.

# 3 Study design

## 3.1 Setup and notation

We use the notation of §2. A task presents a world $w \in W$, the material the output was meant to respect, from which a generating system produces an output y. We write $a(y, w)$ for whether y is acceptable in w and take the corpus's own annotation of $a(y, w)$ as ground truth. An evaluator is a function J that maps whatever it is shown to a report R, a set of claimed defects, each consisting of a span of y and a stated reason. Throughout the paper we manipulate the evaluator's second argument, which is its view of the world. The blind arm computes $R = J(y, \varnothing)$, and the informed arm computes $R = J(y, w)$. J is identical across arms, with the same model, prompt, decoding and rubric, so the arms differ only in whether the world channel is open. Any difference in R can then be attributed to w, up to sampling variation at temperature 0.3, which pooling three votes reduces (Figure 3a). We use these arms in two designs. The paired design scores R against the annotation, whereas the closed-loop design (§3.5) returns R to the generating system and scores the revision that follows.

## 3.2 Datasets

We use seven public datasets, each of which supplies outputs, the world those outputs were meant to respect, and ground truth that requires no annotation on our part (Table 3). We use every dataset as released except IFEval, which supplies prompts but no responses. For IFEval we generated responses with two open models, Llama-3.1-8B-Instruct and Qwen3-14B, and scored them with the corpus's released checker. We use GSM8K in §8 as a second closed-loop corpus, in which correctness is decided by the answer key.

**Table 3:** The seven corpora, the world each supplies, and the ground truth used.

| Corpus | Task | The world w | Ground truth | Used in |
|---|---|---|---|---|
| RAGTruth [25] | QA, data-to-text, summarisation | retrieved passages | marked hallucination segments | §4–§7 |
| WMT-MQM [44] | translation, three language pairs | source sentence | professional error marks by category | §4 |
| FRANK [45] | news summarisation | source article | per-sentence factuality labels | §4, §5, §7 |
| IFEval [46] | instruction following | the instruction | program-checkable constraints | §4, §6, §8 |
| HaluEval [47] | QA, dialogue, summarisation | supporting knowledge | hallucinated/correct pairs | §4–§9 |
| LLM-AggreFact [8] | claim verification, 11 sub-datasets | source document | human support labels | §4, §7 |
| GSM8K [48] | arithmetic word problems | the problem statement | answer key | §8 |

### 3.3 Evaluator conditions

The blind arm receives the output and a rubric naming the defect classes, so it is a strong version of an output-only judge that knows what to look for. What accompanies the output varies with the corpus. On HaluEval the blind arm also receives the question or the dialogue history, and on RAGTruth QA we test adding the question as a separate arm (Appendix A, P31). On IFEval the blind arm of the paired design receives the constraint taxonomy in place of the instruction, because there the instruction is the world. In the IFEval closed loop the blind judge receives the response alone, with neither the instruction nor the taxonomy, and on GSM8K it receives the worked solution without the problem statement, because there the problem is what the solution must respect. In both loops the blind judge therefore lacks the requirements themselves. This is a stricter condition than a deployed judge that sees the user's prompt but not the system prompt or policy. An exploratory IFEval arm that hides only the constraints separates the two (§8.2), and the blind judge of the HaluEval QA loop sees the question and lacks only the source knowledge (§8). The informed arm receives the same input plus w, so the two arms differ only in whether the judge is shown the world. Two controls separate the contribution of the world from that of the judge. The first is a non-generative instrument that has w but no language model (§3.4). The second is a blind language model strengthened with a contradiction detector over its own statements, whose contribution we report in §4. Later sections vary w itself along the dimensions the framework predicts should matter, namely relevance, completeness and form. Each variant is defined where it is used (§5, §6).

### 3.4 Models and inference

The primary judge is Gemini 2.5 Flash, sampled three times at temperature 0.3, with claimed defects pooled across votes in the same way in both arms. Where the judge returns free-text defects (RAGTruth, FRANK, WMT-MQM and the free-text closed-loop reports), we pool by union, so a defect raised in any vote counts. Where it returns a label or a constraint identifier (HaluEval, LLM-AggreFact, the IFEval paired arms and the typed closed-loop reports), we take the majority of the votes that parsed. Union pooling raises recall and false positives together. It applies identically to both arms, but in absolute terms it adds more false positives to the blind judge, which over-flags, and we do not separate that effect. The replication judge is Claude Haiku 4.5 under identical prompts, and the capability ladder in §4 adds Claude Opus 4.8. The non-generative instrument is a DeBERTa-v3-large [49] entailment model fine-tuned for natural-language inference [50]. Its calls are assembled into a typed graph in which requirements, evidence sentences and answer claims form the nodes and support, conflict, coverage and specificity form the edges. Defects are patterns over this graph, and each defect emits a certificate naming the nodes and probabilities that fired. We call this instrument the structural judge and the graph it builds the entailment graph, and both names are used in the figures and in §7. We hold prompts byte-identical across arms except for the block that carries w. All judge prompts, including the closed-loop prompts, are available from the authors on request.

### 3.5 Closed-loop protocol

In the second design (Figure 3b) the report is acted on instead of being scored. R is returned to a responder, which revises its own output, and a program then scores the revision using either IFEval's released constraint checker or GSM8K's answer key. We do not use a judge to grade the repair, because an instrument with the defects that the paired design documents in §4 cannot serve as the outcome measure for revisions made to satisfy it. If the errors of a feedback judge and a scoring judge were correlated, those errors would be registered as improvement. We compare five regulators: the blind judge, the informed judge, an informed judge whose report is restricted to typed constraint identifiers, a perfect certificate emitted by the checker itself, and a control that sends no report beyond an instruction to revise. Unless stated otherwise we report one round of revision, which we extend to three rounds in §8. The responder is the model that wrote the original output: on IFEval, Llama-3.1-8B-Instruct and Qwen3-14B each revise their own responses, and on HaluEval QA and GSM8K the responder is Llama-3.1-8B-Instruct. The feedback judge is Gemini 2.5 Flash throughout, and a regulator that reports no defect takes no action. Intervals on net change come from 2,000 bootstrap resamples of responses, with all the constraints of a resampled response kept together. In the one-round IFEval run the free-text reports were stored with at most 12 defect descriptions per response, and the blind judge's report reached that limit on 397 of 1,080 responses. The three-round run stored the full report and shows the same harm (§8.2).

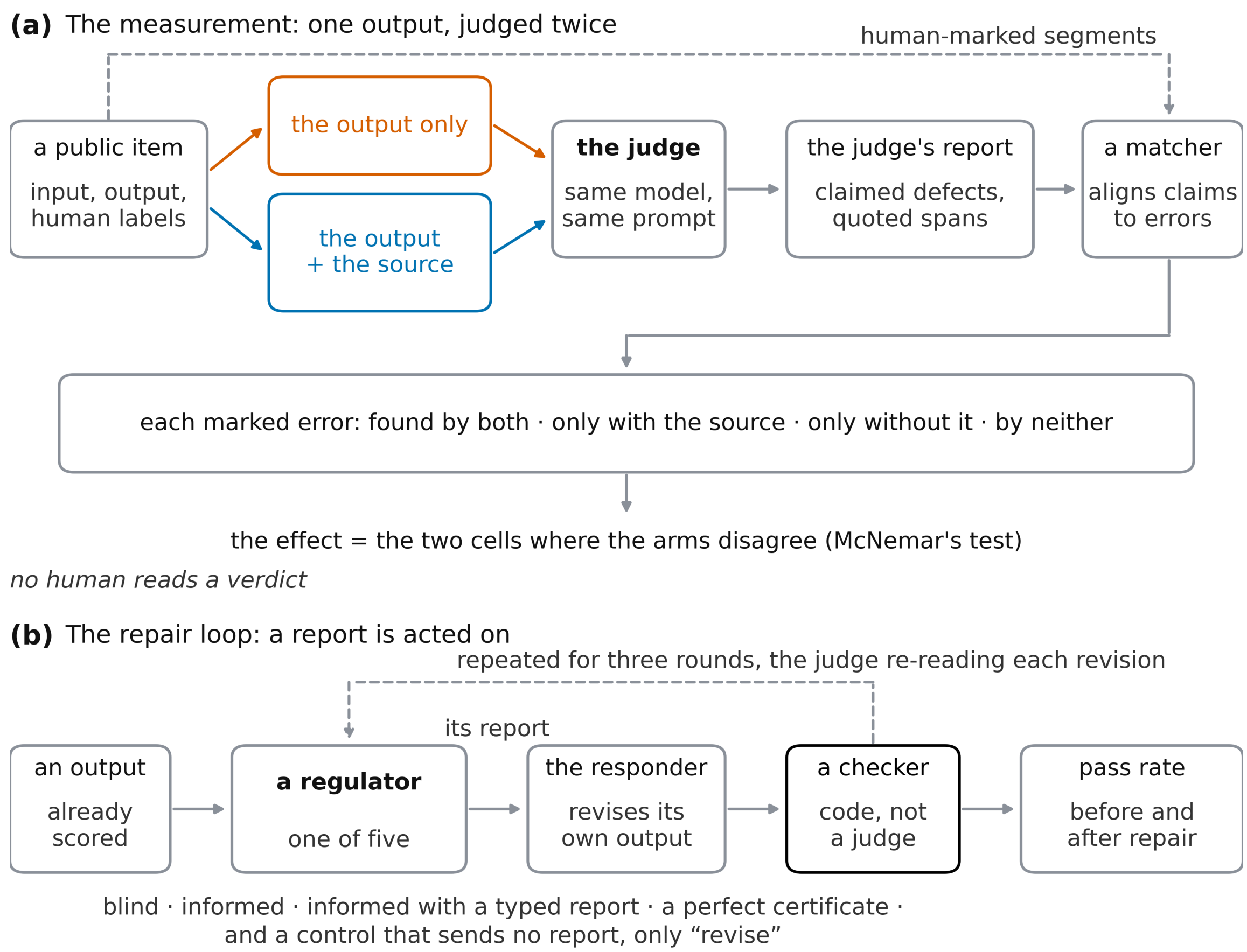

**Figure 3:** How both experiments were run. (a) Paired measurement: each item is judged twice by the same model under the same prompt, differing only in whether the source is attached; a matcher aligns claimed defects to human-marked segments. (b) Closed loop: the report is acted on and a programmatic checker re-scores the revision.

## 3.6 Scoring

How we match a verdict to the ground truth depends on the form the truth takes. Where truth is a label on a sentence, claim or constraint, we compare the verdict to the label directly. Where truth is a marked segment and the judge writes free text, an alignment step is unavoidable. We implement two alignment routes. A strict route credits overlap with quoted text, whereas a generous route lets a small language model decide whether a claimed defect identifies the segment. The generous route is primary, and wherever the two routes differ materially we also report the strict route (§4, §5). On RAGTruth the two routes agree on 70% of blind-arm segments and on 94% of informed-arm segments, so agreement is higher when the judge has seen the world. On the same run, under the generous route the source raises the share of marked errors found from 65% to 91%. Under the strict route the blind arm already scores 95% against 92% for the informed arm, a paired difference of −0.028 ($p = 0.076$), because that route credits the blind judge's heavy quoting without requiring the complaint to locate the error. This asymmetry runs in the same direction as our effect, and we cannot bound its size. The results we lead with in §4 avoid the question entirely, because they use the corpus's own published scoring protocol or the corpus labels. No human reads a verdict in either design.

The following results depend on the generous matcher: the RAGTruth span recalls of §4 to §6 other than the corpus-protocol ladder, the FRANK sentence-level results of §4 and §5, and the WMT-MQM results. The following do not: the LLM-AggreFact and RAGTruth corpus-protocol ladders (P29, P30), the blind-plus-question comparison under the corpus protocol (P31), every IFEval and HaluEval label-scored result, and every closed loop. Our committed protocol excluded manual auditing of matcher decisions, and we performed none. After the committed rounds we audited the matcher mechanically, on 300 blind-arm and 300 informed-arm decisions from FRANK and the same from RAGTruth, in two ways. First, a permuted control paired each review with a marked error from a different item; the matcher credited 0 of 298 such pairs on FRANK and 3 of 300 on RAGTruth, so it does not credit a complaint for being of the right kind. Second, a second model family, Claude Haiku 4.5, re-matched the same real pairs under the same prompt. On RAGTruth it agreed with the stored verdicts on 92% of blind-arm and 96% of informed-arm decisions (Cohen's κ 0.83 and 0.77), crediting 65% against 67% and 91% against 91%. On FRANK it was stricter, crediting 75% against 96% on the blind arm and 72% against 91% on the informed arm, but the difference was of the same size in both arms (21 and 19 points). On FRANK the matcher is lenient but not credulous. Its leniency raises blind and informed recall alike, which works against the gap we report, and on RAGTruth a second model family reproduces its decisions. The audit is exploratory, made after the committed rounds, and is a check by a second model, not by a person (§11).

### 3.7 Statistical analysis

We summarise paired comparisons as 2×2 tables, in which the two cells where the arms disagree carry the entire effect, and test them with McNemar's exact test [51]. We report proportions with Wilson intervals [52] and estimate uncertainty over items by bootstrap resampling of answers rather than segments, because segments within an answer are not independent. We test rank correlations by permutation. We report effects with intervals and p-values as computed, without correction for multiple comparisons, and discuss the consequences in §11.

### 3.8 Internal pre-commitment

We wrote 37 numbered predictions across 17 rounds, each recorded in a file and hash-committed in our private repository before the runs it governs were launched. The prediction ledger, its 17 commit hashes with per-round snapshots, and the verification log are available from the authors on request. The counting unit throughout this paper is the committed clause, so a prediction written with lettered parts counts each part as one unit, and a FAILS clause written for the whole prediction applies to each part. On that rule the ledger holds 83 units, including two that were committed and not run. We take verdicts against the thresholds as written, and no threshold was revised after a result. Each unit receives one of five verdicts: met its bar, fell short without triggering its FAILS clause, triggered its FAILS clause, reversed sign, or not run. Of the 83 units, 32 met their bars, 28 fell short without triggering their FAILS clause, 18 triggered it, and 3 reversed sign with an interval excluding zero (P3, P23 and P32(a)). Two were not run: the Sonnet 5 rung of P1, stopped at 30 of 400 items on cost, and the structural-judge clause of P6. Of the 81 units that ran, the observed direction was as predicted in 50, null in 13 and opposite in 18; a unit whose clauses split in direction is counted by its first-named clause. Table 6 gives every unit with its direction. The prediction commits were made in the authors' private repository and had not been pushed to any remote when this version was prepared. Appendix A lists the round hashes and dates, so they can be checked once the repository is public. Because the commits carry no third-party timestamp, they are internal pre-commitments that readers cannot verify independently, not a pre-registration, and we describe them as such throughout.

We deviated from the committed protocol in five places, each recorded in the verification log. P1: the Sonnet 5 rung was stopped at 30 of the 400 paired-set items when its projected cost exceeded the budget, before the ladder was fixed to the 300-item subset of §4. P10(a) and P10(b): run on the 250-labelled plus 50 error-free prefix subset used for the capability ladder rather than the committed 500 items, with thresholds unchanged and the change recorded before the run completed. P25 to P27: run on a balanced 300-response subset rather than the full 1,080 responses, again with thresholds unchanged and the change recorded before the run completed. P30: the committed protocol instructed judges to quote, but the reused first-phase judgments carried no quote field. A second phase therefore re-judged both arms with one added instruction requesting a verbatim quote field, applied identically to both arms. The first phase, which reused existing judgments, is reported alongside. P36 and P37: a first attempt aborted when the API account ran out of credits, leaving 194 of 300 responses unjudged in at least one arm. That capture is preserved and marked not to be cited. The reported run is a clean re-run from zero rows.

Our scoring departs from the rule of verdicts against the thresholds as written in one place, in the conservative direction. P19(c) clears its bar as written ($\rho = 0.24$ against a bar of 0.15, with no condition on p), but we scored it NOT MET because $p = 0.09$ and the entropy measure sits at its ceiling for most questions. Scored as written, the counts would be 33 met and 27 short of their bars without triggering a FAILS clause.

Some data were collected before the first prediction was committed on 28 August 2026, and none of them is counted in Table 6. The RAGTruth paired run behind the +0.263 recall gain of §4 was complete on 26 August, as were the strengthened-blind control of §4 and an earlier FRANK run of the blind and article arms, scored by the strict route at balanced accuracies of 0.59 and 0.79. P2 replicated the RAGTruth contrast with Claude Haiku 4.5 under a committed bar. P12, committed on 29 August at 01:43 (CEST), was therefore written with the FRANK blind and article results already seen. Its reference arm had not been run, and that run finished at 02:00. P15, committed at 10:38 the same day, was written after the reference-only arm had been seen, and its article-plus-reference arm finished at 10:52.

## 4 Is the channel or the judge the binding constraint? (RQ1)

We measure the deficit on RAGTruth, using the first 250 answers that contain a marked segment (384 segments between them) plus 50 error-free answers. Three judges read every answer, once in the blind condition and once in the informed condition, and were asked to quote the offending text verbatim. We score with the corpus's own character-overlap protocol, ported from the authors' released code, so that no matcher of ours is involved. We then repeated the comparison on 1,100 human-labelled LLM-AggreFact claims, where the judge states whether a claim is supported.

Contrary to our expectation, a more capable judge does not close the gap left by the missing source, and the two matcher-free ladders agree on this. The three judges are Gemini 2.5 Flash, Claude Haiku 4.5 and Claude Opus 4.8. On LLM-AggreFact, scored from the corpus labels with no matcher, their blind balanced accuracies lie between 0.565 and 0.587, and with the document between 0.775 and 0.795 (Figure 4c). Under RAGTruth's own protocol, blind F1 lies between 0.185 and 0.217, and informed F1 between 0.336 and 0.476 (Figure 4a). The frontier judge working blind trails the mid-tier judge working informed by 0.119 [0.092, 0.153] (Figure 4). Blind judges also flag 73–80% of the text of an error-free answer as wrong, whereas informed judges flag 25–57%. Capability is not even monotone within this set of judges, since Gemini's blind F1 exceeds Haiku's.

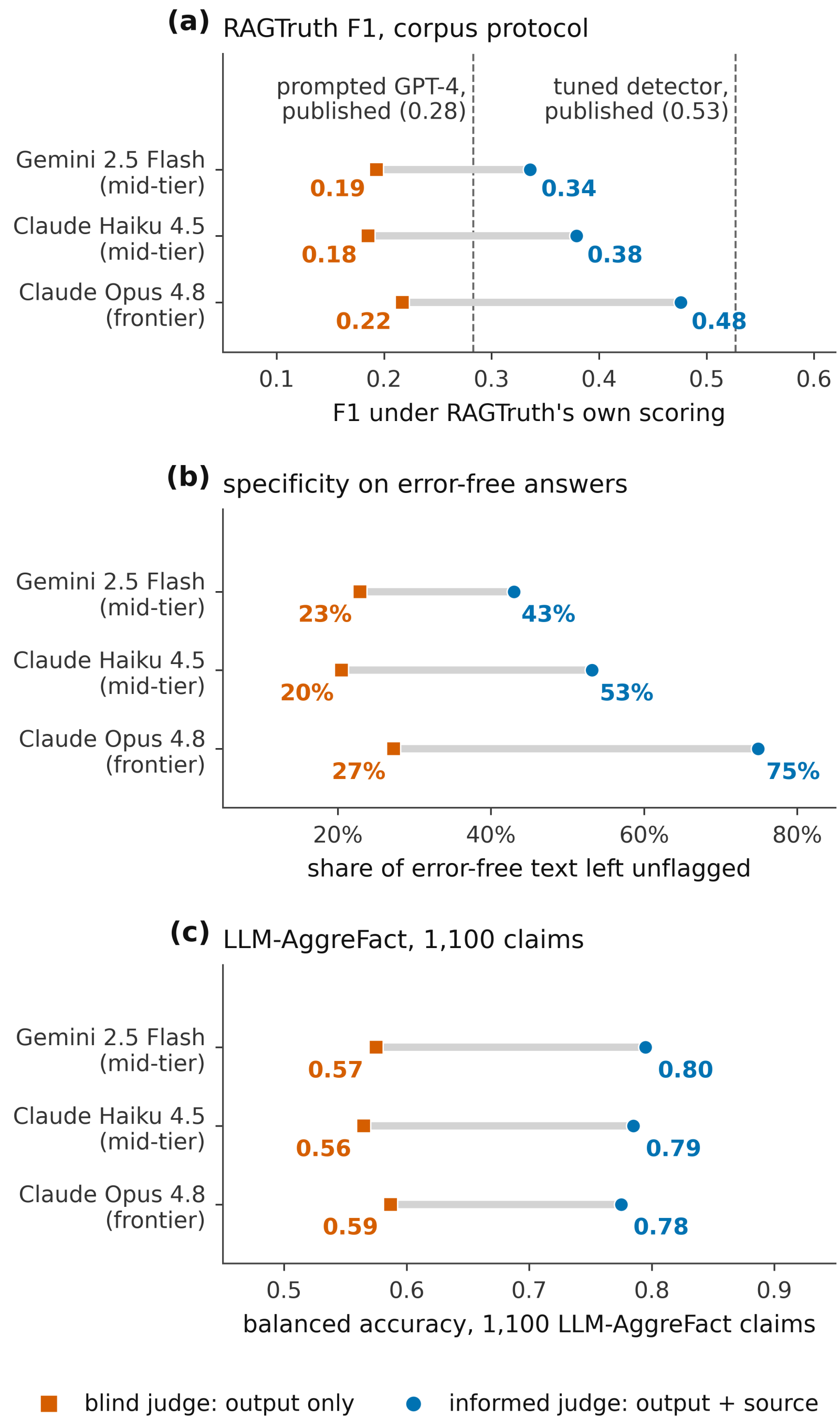

**Figure 4:** A better judge does not make up for not seeing the source. Three judges of increasing capability, blind (squares) and informed (circles), identical items. (a) RAGTruth F1 under the corpus's protocol; published baselines dashed. (b) Character-level specificity on error-free answers. (c) Balanced accuracy, 1,100 LLM-AggreFact claims.

List prices across the three judges span 10–17 times: \$0.30/\$2.50 per million input/output tokens for Gemini 2.5 Flash against \$5.00/\$25.00 for Claude Opus 4.8. Across that span, blind balanced accuracy changes by at most 0.022. As §2 notes, this is a plateau over a narrow range of judges rather than a bound on what capability can recover, but it is the pattern the regulator account predicts, under which a more capable judge still cannot recover a distinction that its input does not carry.

How large the recall gap is depends on how complaints are matched to marked errors. On a larger RAGTruth sample of 400 answers holding 562 human-marked errors, each judged twice, the informed arm under the generous route recovers 173 errors that the blind arm misses and loses 25. It finds 514 errors against 366, about 40% more, a paired gain of +0.263 [0.219, 0.307] with $p < 10^{-27}$ by McNemar's test (Figure 5a). Under the strict route the paired difference is −0.028 ($p = 0.076$), because that route already credits 95% of the blind arm's complaints (Figure 5b, §3.6). We therefore do not rest the claim of this section on the paired gain. Under the generous route the gain holds for both judge families, all three RAGTruth tasks and all six generating models (+0.15 to +0.44). The Gemini run was complete before the first prediction was committed, and P2 replicated it with Claude Haiku 4.5 under a committed bar (+0.227).

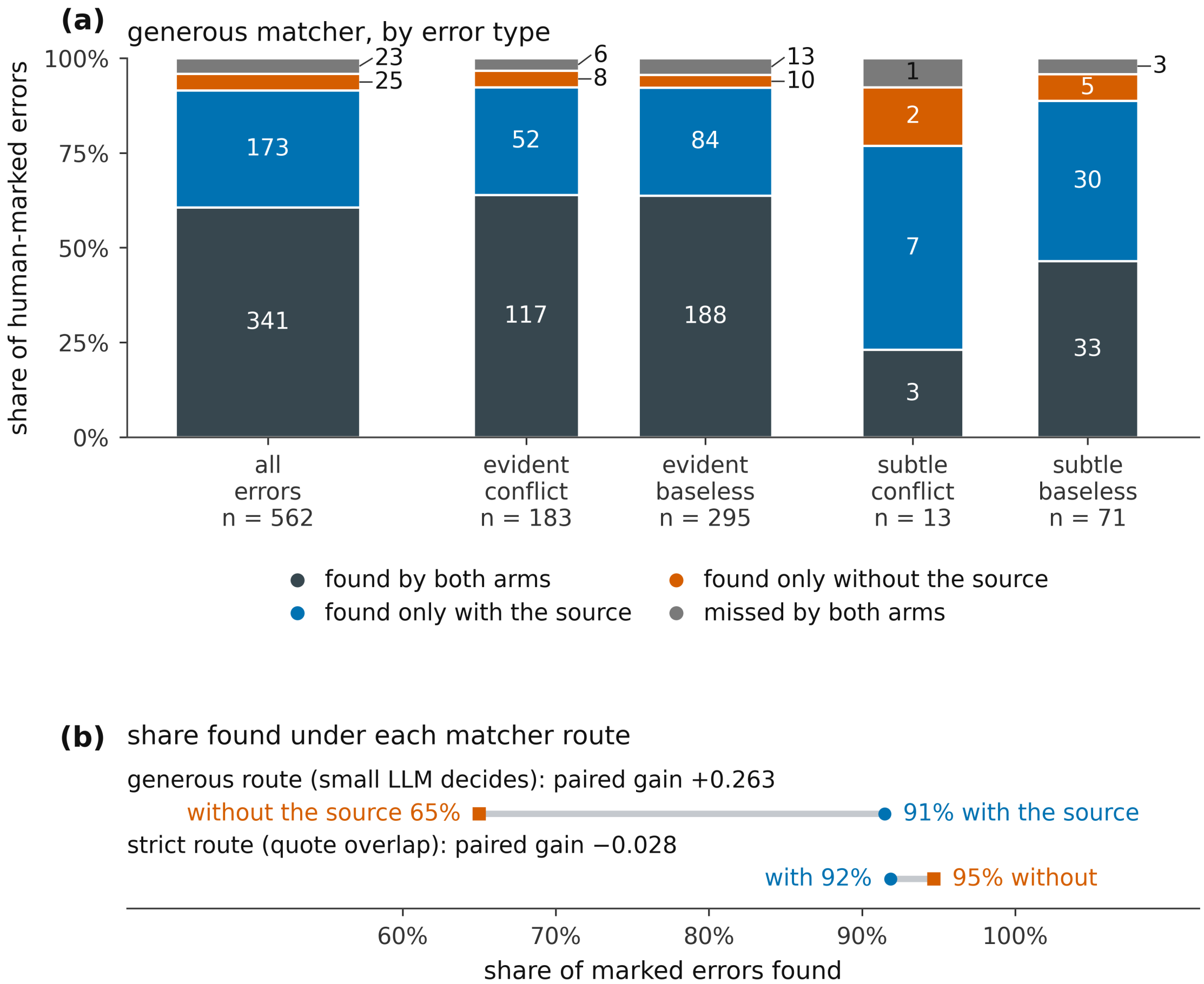

**Figure 5:** Shown the source, the judge finds 40% more errors, but only under the generous matcher. RAGTruth: 562 human-marked errors in 400 answers (a larger sample than the 250-answer ladder of §4), each judged twice by Gemini 2.5 Flash, with and without the source. (a) Generous matcher, by error type. (b) Share of marked errors found under each matcher route; the strict route credits quoted text without requiring the complaint to locate the error (§3.6).

The strengthened blind judge of §3.3 does not close the gap either. On the same 400 answers, adding a contradiction detector over the output's own statements lowers the blind judge's recall from 0.648 to 0.299, and its false positives from 12.4 to 4.3 per error-free answer. Because the marked errors contradict the source while the answers remain consistent with themselves, a consistency check over the output cannot find them. This control was run before the predictions were committed and is not part of Table 6.

In corpora where the blind judge already flags most sentences, the deficit appears as false positives rather than as misses. On FRANK's news summaries the blind judge finds 93% of error sentences but also flags 80% of correct ones, for a balanced accuracy of 0.565, close to chance (Figure 6). When the judge is given the article, balanced accuracy rises to 0.769, and the improvement comes entirely from false positives, which fall from 80% to 31%. The share of errors found falls slightly but significantly (−0.084 on the 310 error sentences both arms scored, McNemar $p = 1\times10^{-4}$). In this setting the world adds no detection and instead stops the judge from flagging correct sentences. These figures come from the runs of 29 August 2026 on 632 sentences, scored by the generous route, the same runs that Figure 6 and §5 use. An earlier run of the same judge on the same sentences, on 26 August and scored by the strict quote route, gave 0.59 and 0.79, a difference of about 0.02 in balanced accuracy, which is the run-to-run variation of this judge on this corpus.

Beyond FRANK, the gain from the document is positive in all eleven sub-datasets of 1,100 LLM-AggreFact claims (Figure 7), and false positives on supported claims fall from 43% to 15%. We sample 100 balanced claims per sub-dataset, so comparisons to published detectors [8] are indicative only. The effect is not universal, however, since on HaluEval dialogue the gain is small, from 0.614 to 0.639, and on long-document summarisation smaller still, from 0.689 to 0.700. In those settings the world is present, but the judge appears unable to use it at that length.

The size of the gap also varies predictably. Before scoring, we coded each of 49 annotation categories by how much of a defect's evidence the output channel cannot carry, which turns Ashby's channel claim into a measurable quantity. The code records where the evidence of a defect sits (supply) and ignores what the question asks for (demand). On this view the requirement is a matter of degree, because variety that a regulator cannot receive reappears as variety it cannot regulate. We accordingly expected the paired gain to grow with the code, and it does (Spearman $\rho = 0.646$, permutation $p < 10^{-4}$) (Figure 8). The mean gain is −0.10 where the evidence is entirely in the text and +0.21 where it is entirely outside. The relationship is fragile, however. Without the two out-of-English translation pairs, $\rho = 0.578$, and these pairs are also where the sign reverses, with supplying the source costing 7 to 11 points. We attribute this to the source redirecting the judge's attention to meaning transfer in a language it reads less natively.

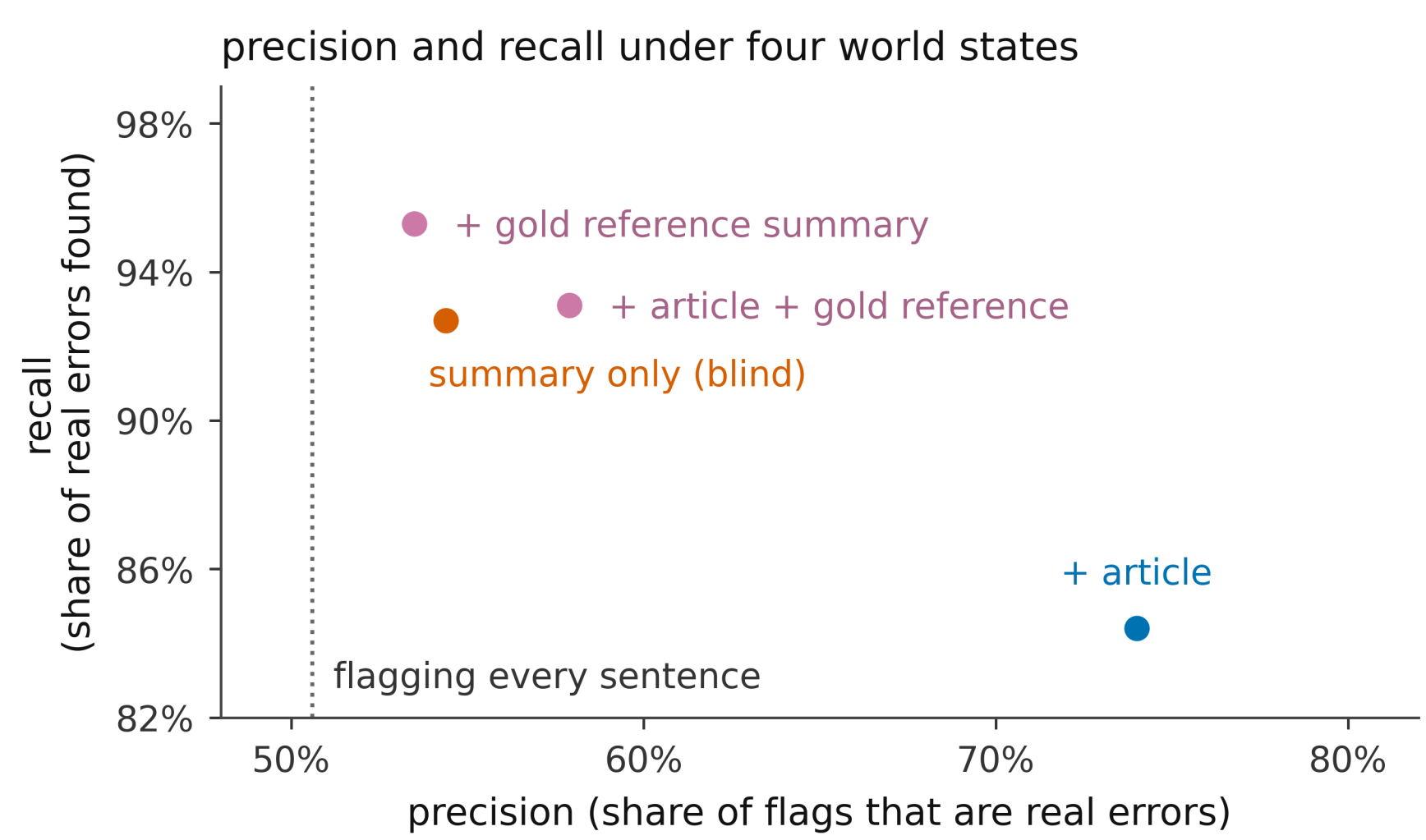

**Figure 6:** Given the article, the judge stops flagging correct sentences. FRANK: 320 error and 312 error-free sentences, the same judge under four world states, all four arms from the runs of 29 August 2026, generous route (P12, P15). Dotted line: the precision of flagging every sentence.

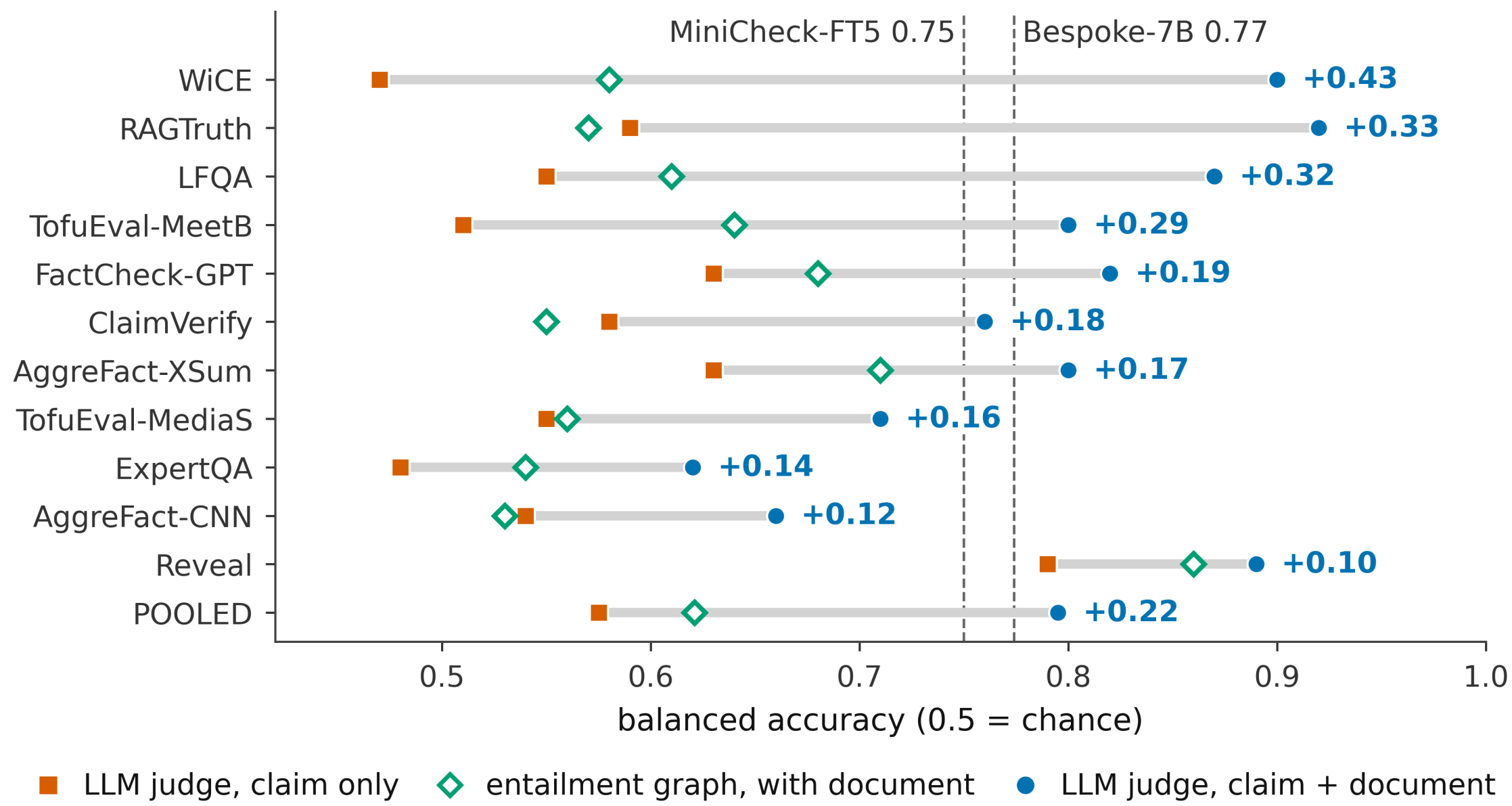

**Figure 7:** Giving the judge the document helps on all eleven benchmarks. LLM-AggreFact, 100 balanced claims per sub-dataset: the judge with the claim alone (squares), the entailment graph (diamonds), and the judge with the document (circles); numbers are the document's gain. Dashed: published tuned fact-checkers on the full test sets [8].

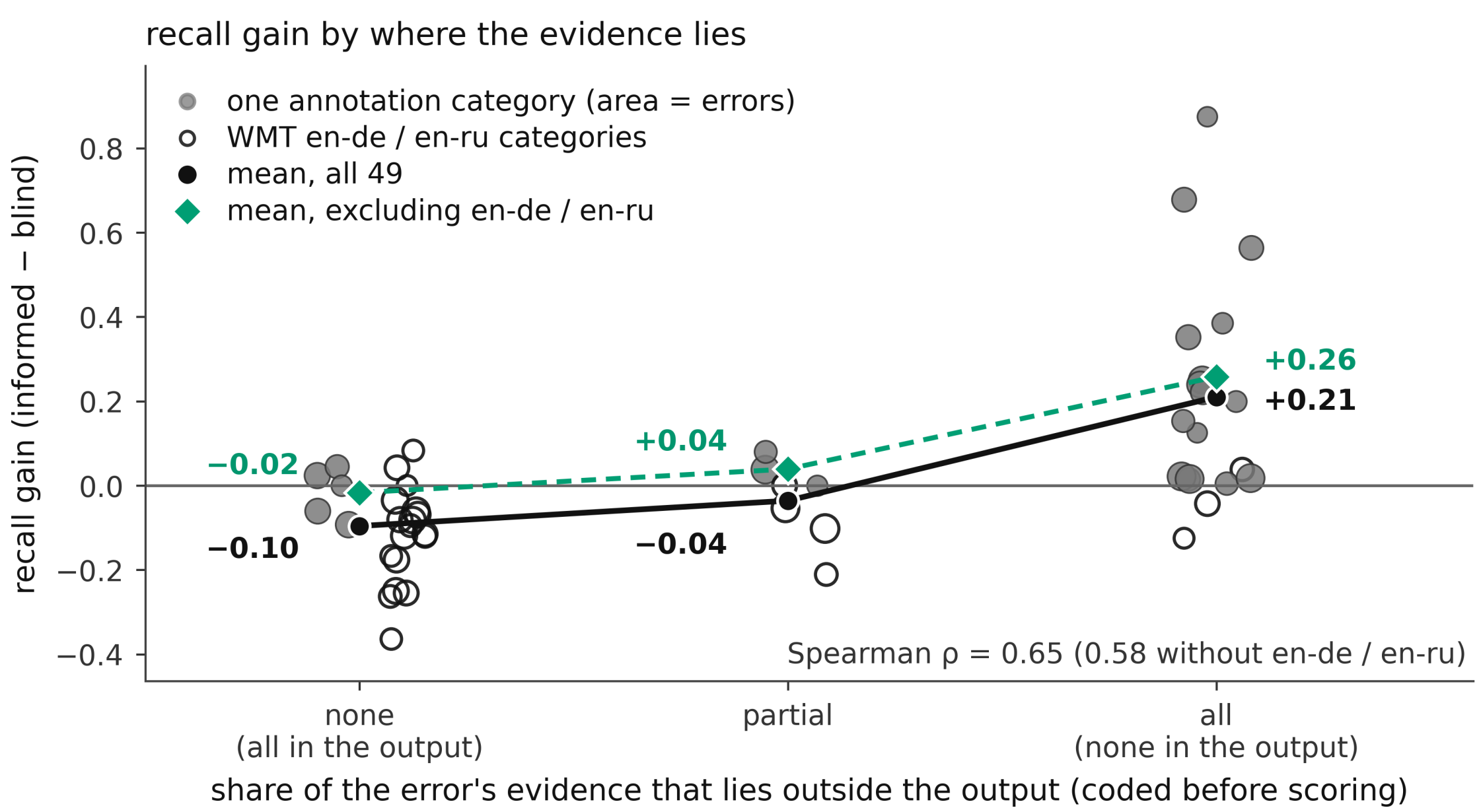

**Figure 8:** The judge gains most on errors the answer alone cannot reveal. 49 annotation categories from four datasets: the three WMT-MQM language pairs, IFEval, RAGTruth and HaluEval. Each category is placed by where its scheme's own definition locates the error's evidence, coded before scoring, against the paired recall gain. Marker area: number of annotated errors.

## 5 Which context helps, and which harms? (RQ2)

A natural reading of §4 is that any additional context helps the judge. We tested this reading with five controls and found that it does not hold, since the control that most resembles help, a gold reference, did the most harm. The first four were each committed as a prediction before its run; the fifth, question demand, was committed as P22 and is reported as directional.

We begin with a gold reference answer, because it most resembles help and was the most informative control. FRANK provides a human-written reference summary for every article, a correct output for the same input. Most evaluation pipelines already have such a reference, and it is the artefact a practitioner is most likely to use. A reference shows the judge an acceptable summary but not what the article said, and, as we describe below, this difference explains the result. The arms in this subsection come from the two runs of 29 August 2026 on the same 310 summaries, scored by the generous route.

When the reference is supplied together with the article, most of the gain from the article is lost. The article alone brings the judge to 0.769 balanced accuracy and 0.739 precision. With the reference added, these fall to 0.618 and 0.579 (Figure 6), so 82% of the precision gain from the article is lost. The loss is on the precision side, because recall even rises (0.844 to 0.931) while false positives more than double, from 31% to 70%. Instead of hiding errors from the judge, the reference leads it to accuse correct sentences, as in the reference-only arm below. The two-input judge also behaves more like the reference-only arm than like the article-only arm, since at 70% false positives it sits 15 points of false-positive rate from the reference-only arm and 39 points from the article-only arm. When both channels are open, the judge appears to rely mainly on the reference.

Used alone, the reference is no better than the blind judge. A judge given the reference instead of the article scores 0.552, against 0.565 for the blind judge (Figure 6). Its precision is 0.535 against 0.544, and false positives move from 80% of correct sentences to 85%. McNemar's exact test on the paired flags over the error-free sentences gives $p = 0.133$ on the generous route, within noise, and $p = 0.047$ on the strict route, so the rise is not robust to the scoring route. The ordering of balanced accuracies is the same on both routes (0.562 against 0.574 on the strict route). We believe the explanation is that a reference is only one of many acceptable summaries, which legitimately differ in what they select. A faithful sentence that the reference happens not to cover has no support in the reference, so the judge reports it as unsupported, even though the article would have supported it.

One possible explanation for why the reference dominates is the relative effort of the two comparisons. Grading against a reference is a sentence-to-sentence comparison between texts of the same shape and length, whereas grading against an article requires the judge first to locate the relevant passage, which may not exist. Error control is inexpensive and always available, whereas cause control costs more and sometimes fails to find its evidence. A judge offered both appears to take error control. We stress that this last step is our interpretation of the mechanism and not a separate measurement. What we measure is that the false-positive behaviour with both inputs resembles the reference arm and not the article arm.

This distinction corresponds to the one Ashby draws (§1). A gold answer supplies the goal set without the disturbance channel, so regulation against it is error-controlled and can at best be partial. Our measurement adds two observations that Ashby's account does not contain: the reference is weaker than the article and on its own lowers the score, and when both are supplied, the reference, as the error-controlled comparator, dominates the article, the channel that carries the disturbance. Both observations were committed as predictions in advance (P12 for the reference against the article, and P15 for the two together), and both held. The four arms come from one run family on one day, so the comparison is internally controlled. Two further runs of the two reference arms, with fresh judge votes, reproduced it: the reference alone scored 0.552, 0.555 and 0.551, and the reference beside the article 0.618, 0.619 and 0.627. The result is also scoped to one corpus and one judge, on a task where many outputs are acceptable. Where a task has a single correct answer and a program compares the output with it, as with the GSM8K answer key in §8, that comparison is the most precise regulator we measured. What fails on FRANK is handing a judge one acceptable output as its evidence, not the use of a known answer by a checker. An exploratory run on HaluEval QA, where each question has one correct answer, repeated three times with fresh judge votes, shows that this scope matters. On the same 500 pairs, a judge given the reference answer in place of the knowledge reached a balanced accuracy of 0.919, against 0.852 with the knowledge and 0.645 with neither, and adding the reference to the knowledge raised balanced accuracy by 0.051 [0.033, 0.067]. The two further runs gave 0.922 and 0.920 for the reference alone and +0.051 and +0.048 for adding it. Part of the reference arm's advantage is mechanical, since a correct answer under review repeats the reference, which makes its low false-positive rate (0.044) easy to reach. Its recall on hallucinated answers, 0.882 against 0.850 with the knowledge, is the fairer comparison. We conclude that a reference helps where it fixes the one acceptable answer and harms where it is one acceptable output among many.

The second control replaced the true source with a placebo, namely another item's source of the same task and length. Recall rose above its blind level, from 0.607 to 0.761, and false positives rose from 12.4, the blind rate on the full run's 100 error-free answers, to 20.9 per error-free response, a relative rise of 69% (Figure 9). The true source, by comparison, produced 7.5 false positives per error-free response at a recall of 0.923. The judge flagged more text, and some of the extra flags happened to land on marked errors. We refer to this behaviour, flagging without supporting evidence, as accusation. We read it as variety without a model, since a placebo drawn from another item carries no information about the errors in the response being judged. The same manipulation has a larger effect on the generating side, where an agent given another item's knowledge answers with accuracy 0.20, below the 0.32 and 0.45 that the two agents score with no knowledge at all (§9).

The third control concerns the judge's prior belief. On HaluEval QA we predicted that removing the question would hurt the blind judge, but balanced accuracy instead improved by 0.145. With the question, the blind judge flagged 54% of correct answers, against 20% with the answer alone and 15% with the knowledge. The question supplied no evidence about the answer, but it appears to have elicited what the judge believed the answer to be, and this belief then acted as an internal gold reference. The two blind arms differ in prompt wording as well as in the presence of the question, so we cannot isolate the mechanism.

For the fourth control, dilution, we buried the true source among 2, 5 or 10 irrelevant documents, which cost 2 to 5 points of recall. At this scale dilution has a small cost, and relevance matters much more than volume.

Finally, we had a language model annotate each RAGTruth question, from the question alone, following [53], according to whether a complete answer demands every answer or accepts any one. For questions that demand every answer, the blind judge misses more (recall 0.534 against 0.705) and the world adds more (+0.389 against +0.251). The difference of +0.138 has an interval that crosses zero and $p = 0.052$, so we report this result as directional and underpowered (Figure 10).

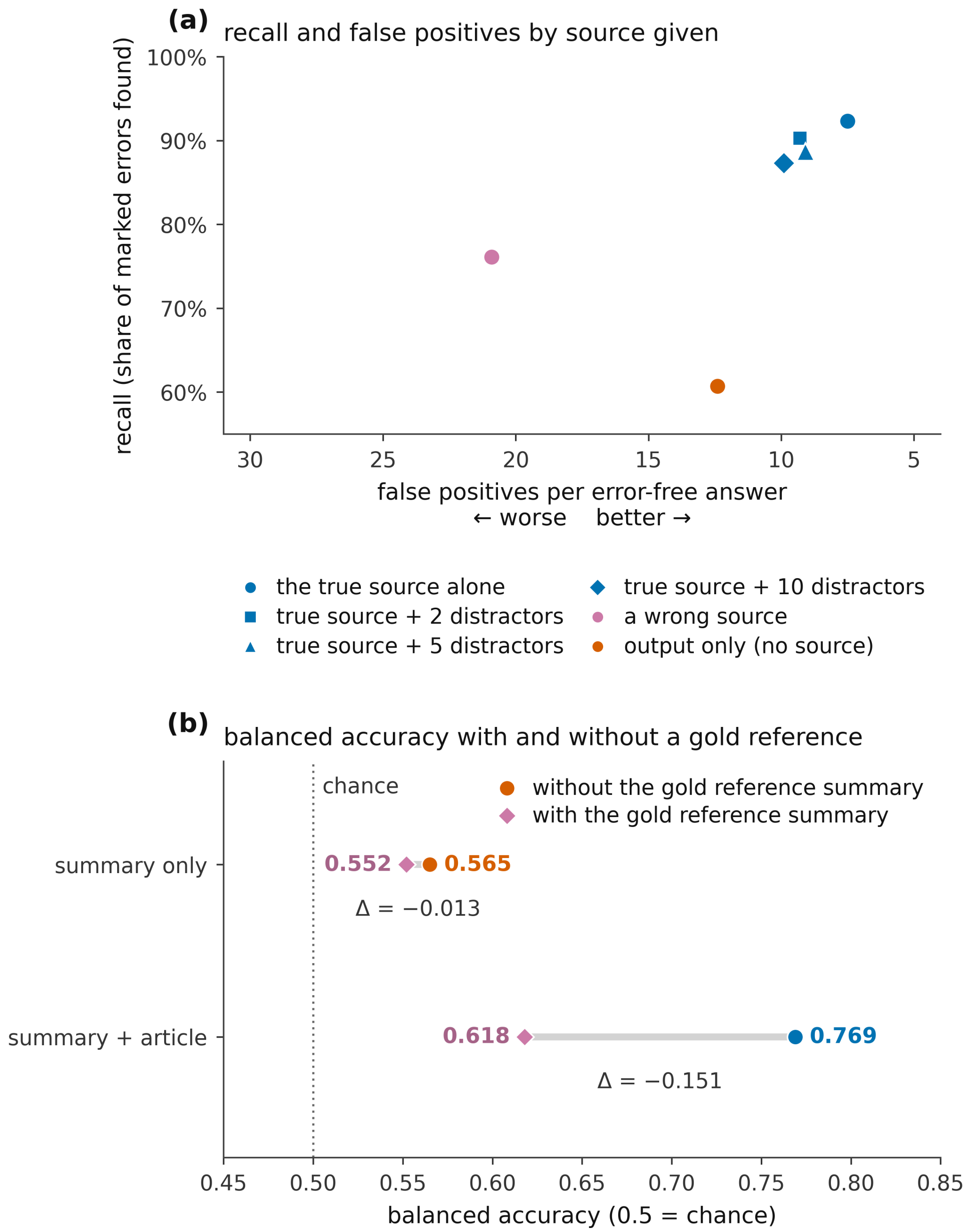

**Figure 9:** Judge performance depends on which source the judge is given. (a) RAGTruth QA, the first 250 answers with errors and 50 error-free answers: the same judge under a wrong source, the true source, and the true source among 2, 5, and 10 distractors. (b) FRANK: the four arms of Figure 6 as balanced accuracy, with and without a gold reference answer, without and with the article. RAGTruth counts false positives only on its error-free answers, so panel (a) plots false positives per error-free answer rather than precision.

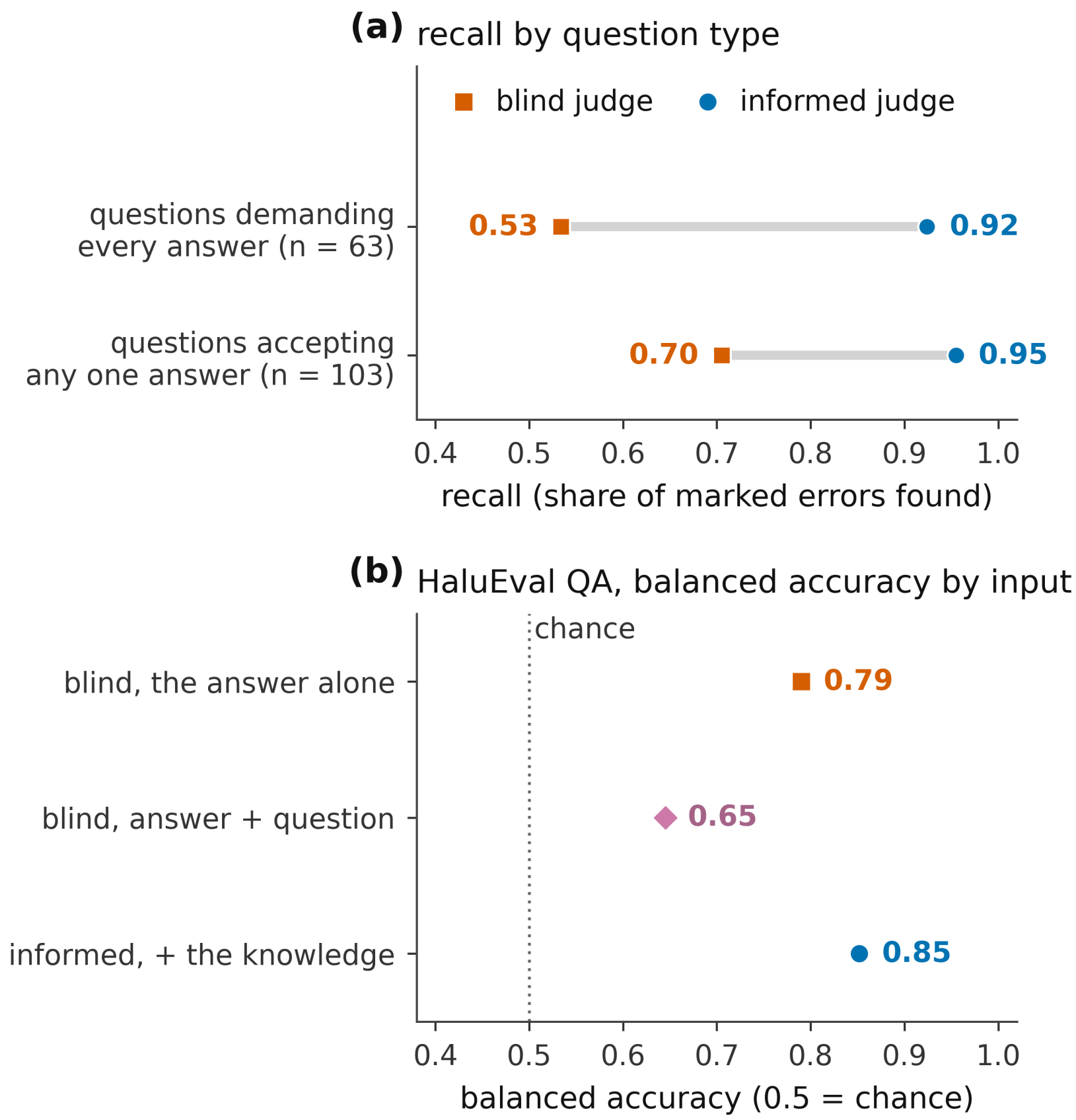

**Figure 10:** Questions demanding a complete answer are where the judge fails without the source. (a) RAGTruth QA, 166 answers, with each question annotated, from the question alone, for whether a complete answer demands every answer or accepts any one; recall by stratum. (b) HaluEval QA, 1,000 items, balanced accuracy: the answer alone, plus the question, plus the knowledge.

## 6 How much must arrive, and in what form? (RQ3)

Supplying the right evidence is only part of the problem, because how that evidence is delivered to the judge also matters. To measure delivery, we ran a $2^4$ factorial on 166 RAGTruth answers that varied four factors at once: the question shown, the source shown, typed requirements, and the instrument. The factorial lets us measure each factor's contribution as its exact Shapley value [54], rather than adding conditions in one arbitrary order. Figure 11 shows the design and Figure 12 the result.

The factors turn out to do different regulatory work, and Ashby's account distinguishes the two kinds. No evaluator can match the full variety of a language model's output, because that variety is unbounded, but a regulator can attenuate the variety that reaches its essential variables and then match what remains. The factorial separates these two operations. Typed requirements do the attenuating, since they fix what counts as a defect and remove false positives without finding additional errors. The source does the matching by supplying the variety that survives attenuation. For the language-model judge, a decomposition over three factors (the question, the source and typed requirements) gives the source the whole recall gain (+0.322), while typed requirements remove the most false positives (Figure 12). The committed prediction, P14(a), was stated over all four factors including the instrument, and it failed. In the four-factor decomposition the instrument receives the larger recall share (+0.181 against +0.105 for the source), because the entailment graph without a source calls every claim unsupported and so reaches full recall, at 9.45 false positives per error-free answer (Table 6). The three-factor result is therefore a post-hoc decomposition within the language-model judge. The shown evidence governed detection, and the instructions governed reporting. The practical rule is therefore to attenuate the requirements and leave the evidence unattenuated. A 25-type taxonomy and a pre-routed evidence set both attenuate the wrong side, and both failed. On IFEval, the taxonomy gained 7.7 points of recall at the cost of 19 points of false-alarm rate. On HaluEval summarisation, routing changed balanced accuracy from 0.692 to 0.680.

The three source sentences nearest each claim make up 11% of the source, yet they carry almost all of the variety the judge needs to match. Shown only these sentences, the judge keeps 98% of the recall it gets from the whole source, 0.908 against 0.927 (Figure 13). Its false positives are higher, however, 3.73 against 3.36 per error-free answer, so this test failed the bar we had committed to in advance. The complement (the source without those sentences) nearly triples false positives.

How the source is presented also matters. Numbering the source sentences and requiring a citation for each defect reduced false positives from 3.36 to 2.88 at a cost of 0.5 points of recall, the least expensive precision gain in the paper. Giving the language model the structural judge's evidence graph as text had the opposite effect, raising recall to 0.979 at 5.5 times the false positives of the full source. The graph's format transferred to the language model, but its discrimination did not.

We also asked the judge to restate the world before judging (the restate-then-judge test, P32 in Table 6), a natural repair for the long-document failure in §4, and found that it made the judge worse. On HaluEval summarisation, accuracy fell from 0.700 to 0.665 and false positives rose, even though the restatements were largely faithful (median fidelity 0.90 by entailment). A token-matched control that restated the rubric instead performed worse still. A blind judge given the same restatement step under-flagged instead of over-flagging as the placebo-fed judge of §5 did, and its recall fell from 0.83 to 0.65, consistent with the judge confirming its own reconstruction.

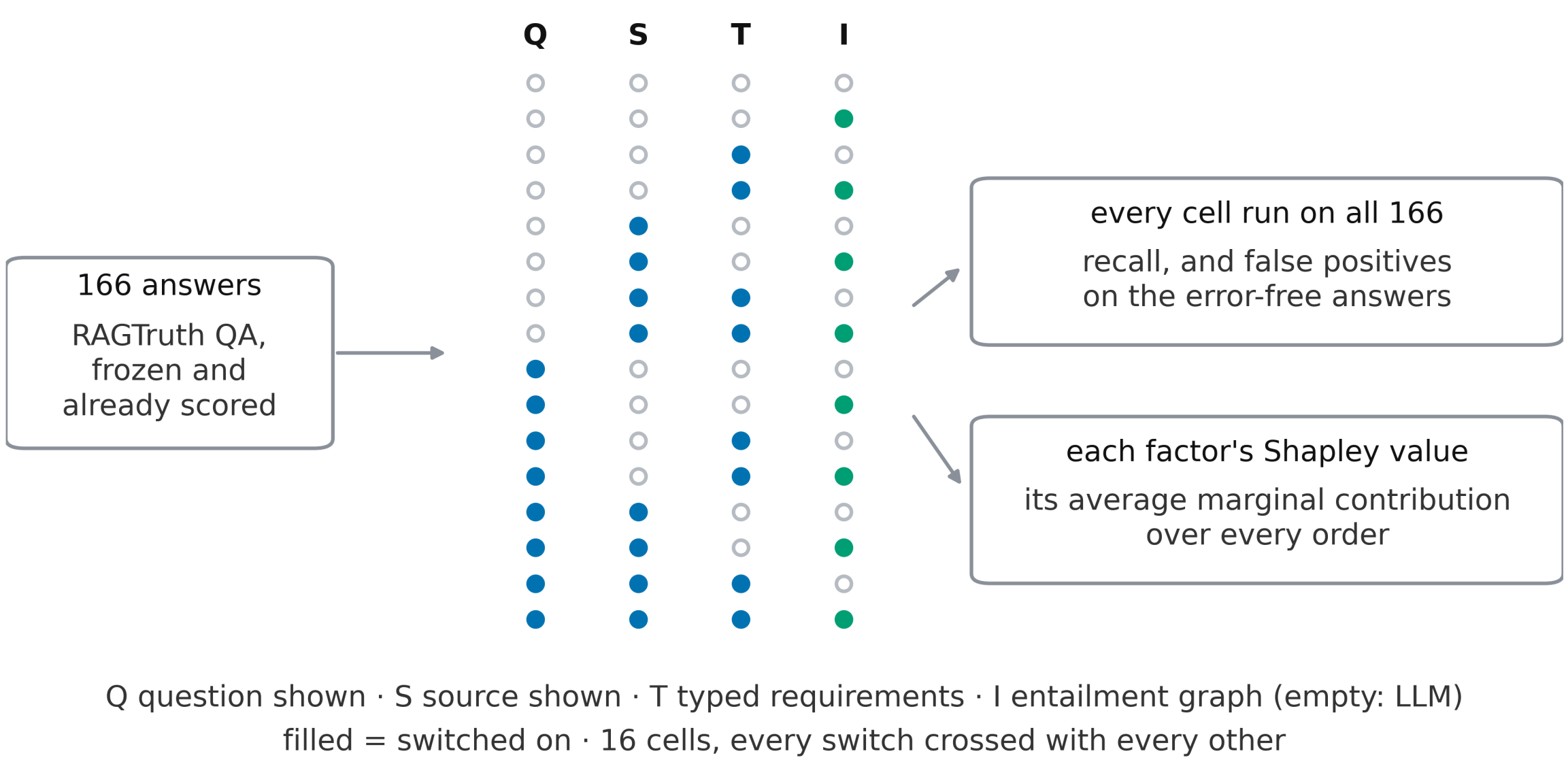

**Figure 11:** How the sixteen cells were built. The factorial: the question, the source, typed requirements, and the instrument, crossed on the same 166 answers; every cell run.

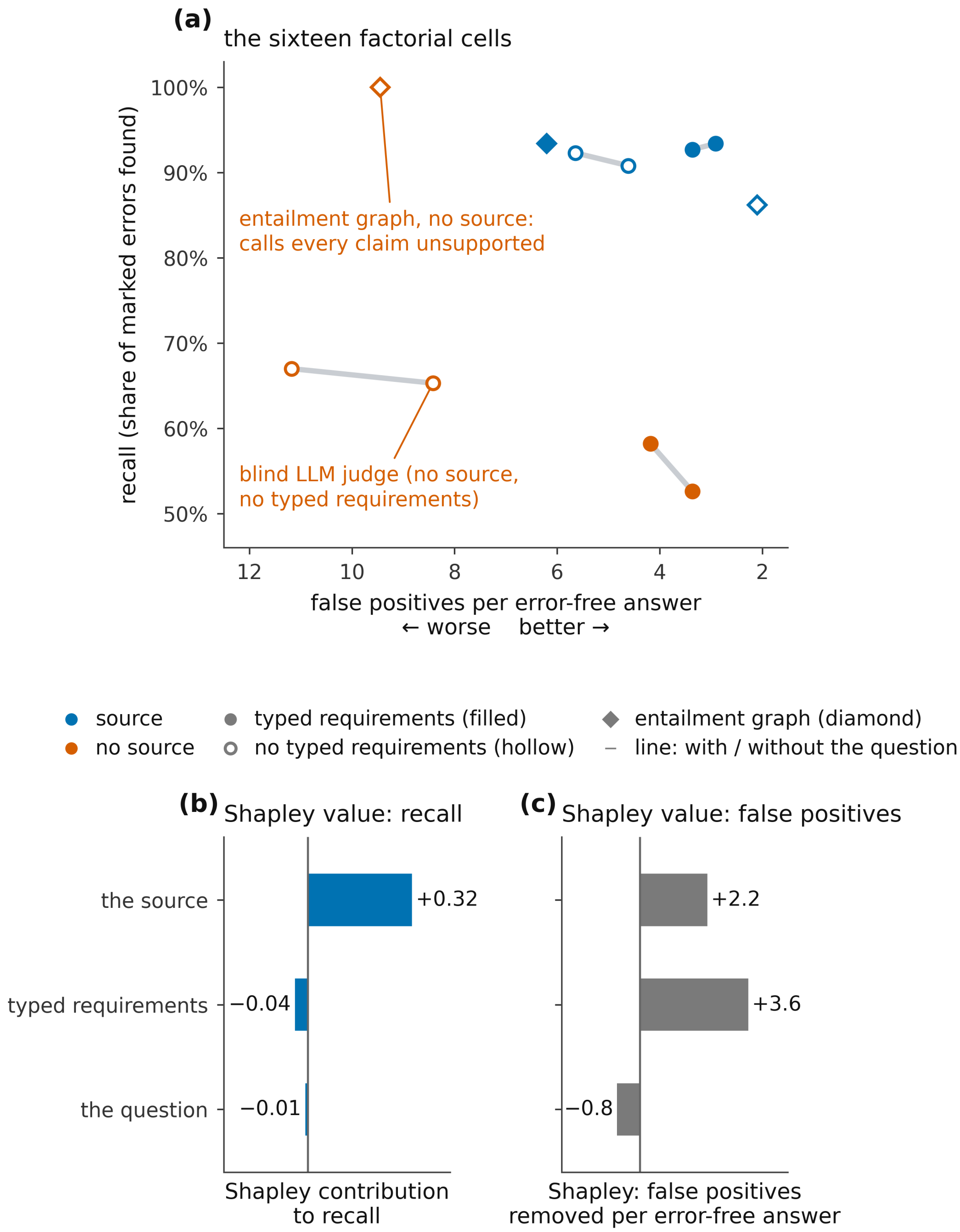

**Figure 12:** The source decides what the judge finds; the requirements decide what it reports. RAGTruth QA, 166 answers. (a) The measured cells: recall against false positives per error-free answer. Filled marks: typed requirements; diamonds: the entailment graph. (b, c) Each factor's Shapley contribution to recall and to false positives removed.

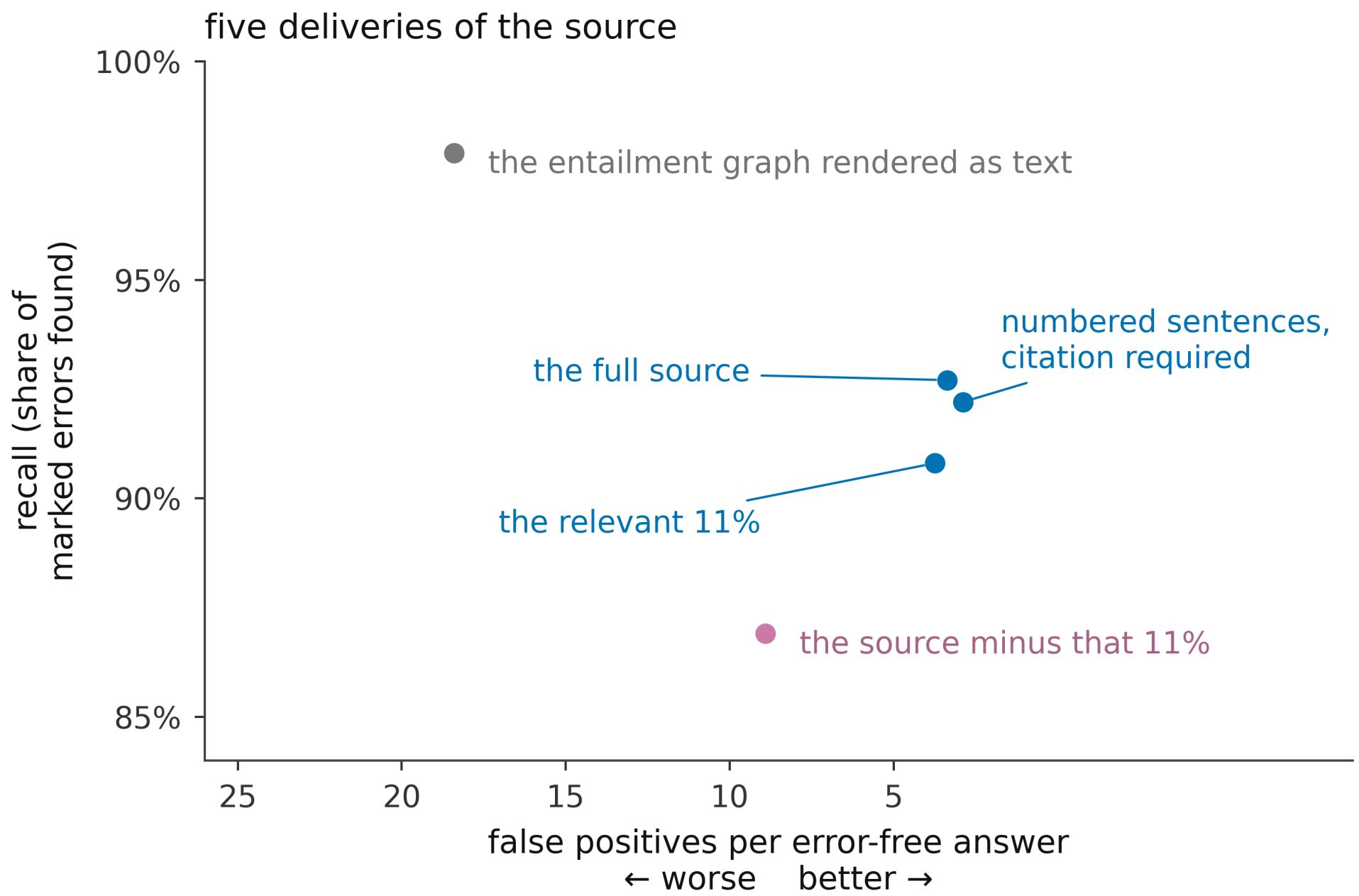

**Figure 13:** Three sentences do the work of the whole source. Five deliveries of the same source on the same 166 answers, to the language-model judge shown the question and typed requirements: the full source, the relevant 11%, the source minus that 11%, numbered sentences with a citation required, and the entailment graph rendered as text.

## 7 Must the evaluator be a language model? (RQ4)

Whereas the preceding sections varied what the judge sees, in this section we hold the world fixed and change the regulator itself, the instrument that does the judging. The structural judge introduced in §3 uses no generative model. Following work on entailment trees [55] and typed entailment rules [56], it assembles the calls of a DeBERTa entailment model into a typed graph and reads defects as patterns over that graph.

On grounded text, the advantage of the language model as a regulator does not come from generation. Figure 14 shows the comparison by task, in which the graph is better on data-to-text and the language model is better on question answering. Pooled over 390 answers, the graph reaches a recall of 0.927 and the language model 0.915, while the language model raises 7.5 false positives where the graph raises 5.6. On detection the two instruments are at parity, and the graph, a regulator with no generative model in it, produces a quarter fewer false positives.

This gain comes from the structure of the regulator, since the entailment model inside it performs poorly when used alone. A plain pairwise use of the same entailment model reached a recall of 0.64 on RAGTruth, and re-engineering it did not raise this ceiling. The entailment model scores claims that contain a marked hallucination as entailed, with a median of 0.986 against 0.995 for error-free claims, so the instrument is confidently wrong. Assembling its calls into a graph improved the result where tuning could not.

The structural judge, which matched the language model on grounded text, fails when the task involves paraphrase. On 200 HaluEval summarisation pairs it flags 96% of the hallucinated summaries but also 92% of the correct ones, so it flags almost everything and reaches a balanced accuracy of 0.522. On LLM-AggreFact it reaches 0.621 against the published 0.750 of MiniCheck-FT5 (Flan-T5-L) [8], and so fails the bar we had committed to. An abstractive claim paraphrases its source, and our zero-shot entailment model treats paraphrase as unsupported. On FRANK the ordering predicted by evidence location holds, with the structural judge's recall at 0.83 to 0.95 on the article-evidence categories and 0.77 on grammatical errors. The one exception is coreference, at 0.778, which the theory places higher. The grammar figure comes from a grammar-enriched sample of 113 additional summaries rather than from the random sample.

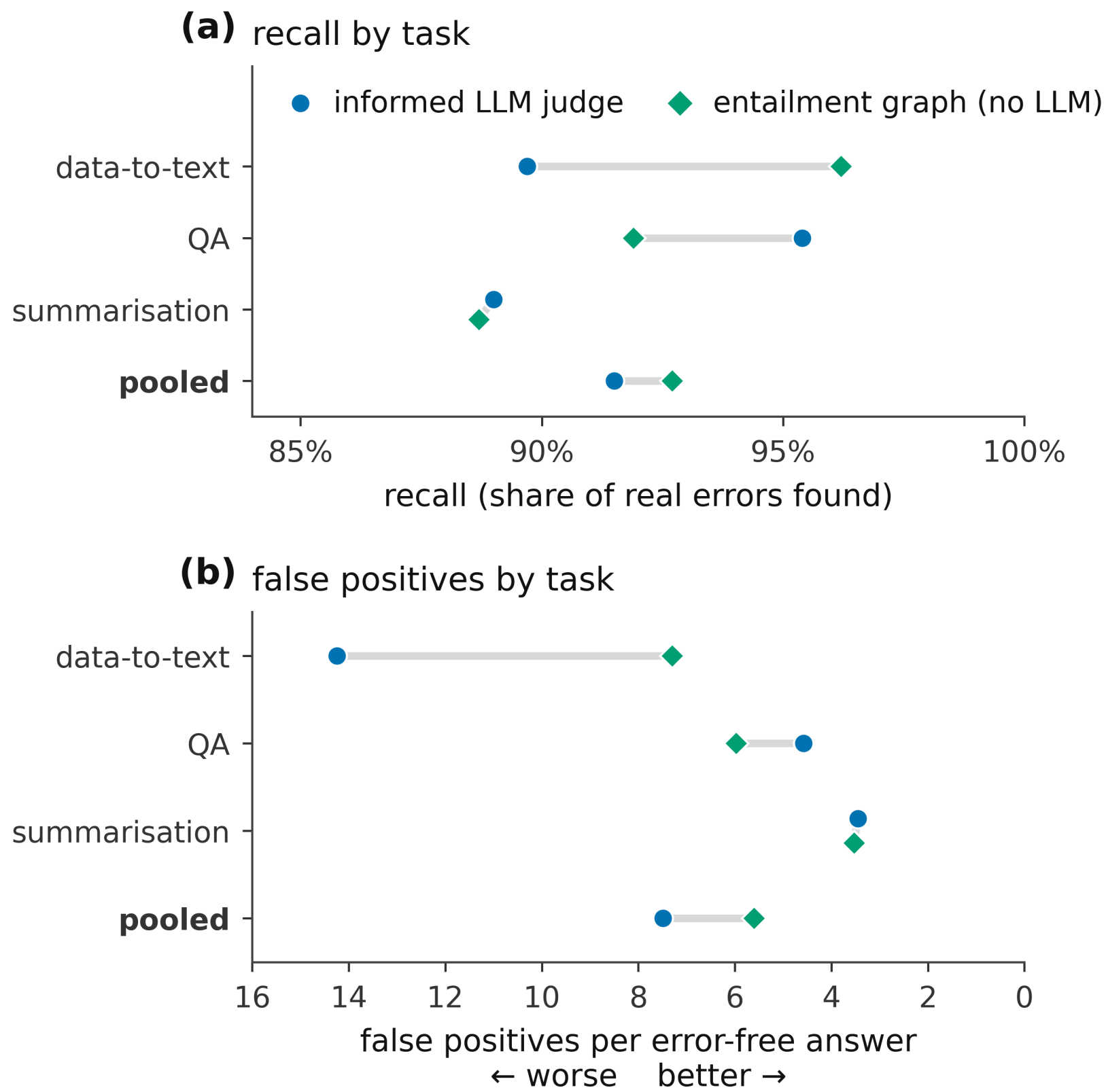

**Figure 14:** A small entailment model finds as much as the language model does. RAGTruth, 390 answers, by task, rows shared across panels: the entailment graph (diamonds) against the informed language model (circles). (a) Recall. (b) False positives per error-free answer; axis reversed so better is right in both panels.

# 8 When do a judge's reports do harm? (RQ5)

## 8.1 One repair round

We turn to the closed loop, the result that matters most for deployment. When its reports are acted on, a judge that cannot see the world leaves the system below its starting point on IFEval and HaluEval, while on GSM8K the loss is within noise, with an interval reaching zero. On IFEval alone it leaves the system below the no-report control. Whereas the preceding sections measure what an evaluator can see, in a deployed system an evaluator's reports are acted on as edits, retries and escalations, which makes the evaluator a regulator. IFEval lets us close this loop without circularity, because a checker scores the repair rather than a judge. Five regulator arms were run on 1,080 responses covering 1,664 constraints: a blind judge shown only the response, an informed judge shown the instruction, an informed judge restricted to typed constraint ids, the certificate, and a no-report control. The responding model revised its own output, and the checker re-scored the result. In the no-report control the responder is told only to revise, which separates the harm a report causes from the harm caused by revision alone and sets the baseline against which "worse than nothing" is defined. Because the blind judge sees neither the instruction nor the constraint taxonomy, the IFEval result concerns a judge missing the requirements altogether, not a judge that sees the user's request but not a separate system prompt. §8.2 tests that configuration separately, and hiding only the constraints does nearly the same damage.

On IFEval, we find that a blind judge's reports leave the system worse off than no report at all. The no-report control gains +0.025, and the reason for this gain matters, because "no report" does not mean "no information". The responder still holds the instruction, so the control removes only the judge's verdict and leaves the world in place. In a second pass over its own answer, with the requirements still available, the responder fixes 98 constraints it had missed and breaks 56 that were already satisfied. Revision has a cost, since it disturbs correct work, but on IFEval it fixes more than it breaks. On the full 1,080 responses the blind judge changes the pass rate by −0.145 [−0.168, −0.123], 17 points below the no-report control, breaking 331 constraints while fixing 90 (Figure 15).

Comparing the two arms isolates what the report itself contributes. The report does not help the model fix more constraints (90 against 98 for the no-report control), but it multiplies the number broken by six (331 against 56). In an unplanned follow-up analysis, which was not a committed prediction, we examined the 331 broken constraints. Of these, 185 were constraints the judge had flagged although they were satisfied, and the other 146 were constraints it had not mentioned, broken as a side effect of edits aimed elsewhere. The informed judge gains +0.038 and the certificate +0.059. The informed judge's margin over the no-report control is small (+0.013 on the full corpus, and within noise on the 300-response subset), and on that subset the no-report control's first-round gain of +0.059 exceeds the informed judge's +0.048. Because free text and the no-report control also tie at +0.059 in the channel experiment of §8.2, we do not claim that the informed judge beats the no-report control on IFEval. Access to the world is not enough for a regulator to produce a net gain in a loop, because the gain depends on the precision of its report, which we measure directly in the next subsection.

### 8.2 What governs the harm

We ran the loop for three rounds, with fresh judging in each round. Under the blind judge the pass rate falls every round, from 0.851 to 0.731 after the first round, 0.706 after the second and 0.683 after the third (Figure 16). The informed judge converges, gaining +0.048 in the first round, +0.025 in the second and +0.002 in the third, and the certificate saturates at 0.931. Two variants of the informed judge did no better than the no-report control, which ended at 0.899: sampling the judge once instead of three times ended at 0.874, and acting only on defects that both the free-text and the typed-id judges reported ended at 0.893.

While the number of constraints fixed is similar across regulators (90 to 128), the regulators differ in what they break, and this follows the accuracy of their reports. To test this directly we varied the report channel while holding the judge fixed, using the 300-response subset described in §3.8, where the no-report control gains +0.059. A bare 1–5 score gives +0.036, below the no-report control. A score on a 1–5 scale carries at most $\log_2 5 \approx 2.32$ bits, and the median of three votes is again one of the five values. The scores the judge emitted have an entropy of 1.56 bits, since 192 of the 298 responses it scored received a 5 (a descriptive statistic, not a committed quantity). Free text gives +0.059, typed constraint ids +0.013 and the certificate +0.074. On satisfied constraints, the false-positive rate is 0.012 for the certificate, whose report names only constraints the checker had failed, so what its arm breaks is collateral of revision, 0.519 for free text, and 0.605 for typed ids. A scalar names no constraint, so no rate is defined for it. The two ends of this ordering follow precision, with the most precise channel gaining most and the least precise gaining least. The ordering does not follow the number of bits carried, since the certificate emits the fewest distinct reports of any defect list and gains the most.

We conclude that Ashby's capacity bound is necessary but far from sufficient. This ordering by precision was not among the predictions committed for IFEval, where the committed prediction, an ordering by regulator, was P24(a) and failed. We read the precision ordering from the P24 data after that failure, then committed it as P35(c) and confirmed it on HaluEval and GSM8K. Each of those orderings has three arms, one of them the oracle, so a perfect rank correlation has a permutation p of 1/6 on each corpus, and because the oracle is perfectly precise and never breaks a correct answer, part of each ordering holds by construction. We therefore read the precision account as a description consistent with every loop we ran, not as a confirmed prediction. Two later committed predictions support the same reading: the report-channel ordering (P36(c)) and the stated-goal precision result (P37(b)).

We also swept the variety of the world and the gain of the loop. We showed the judge none, a random subset (63% of the constraints as realised; see §11) or all of an instruction's constraints, and the loop moves from −0.120 to +0.042 to +0.048 (Figure 17). Most of the gap closes well before the world is complete. Throttling the gain, by acting only on the blind judge's most severe complaint instead of all of them, reduces the harm by about 60%, from −0.101 to −0.038, but does not remove it. The two sweeps are separate runs with fresh judging, which is why the blind arm reads −0.120 in one and −0.101 in the other. No gain setting we tested made the blind judge, a regulator with no model of the world, safe.

Stating the goal to the judge, by asking whether the response satisfies every constraint in the instruction instead of whether it is good, did not improve the net outcome (+0.055 against +0.065, with an interval crossing zero). It did reduce the judge's false-positive rate on satisfied constraints, from 0.553 to 0.405, and narrowed the set of responses acted on, from 275 to 228.

The pattern replicates on two further corpora, where it can also be separated from the effect of revision pressure. On HaluEval QA and GSM8K the regulators that use a judge again fall in order of report precision, and the certificate broke no correct answers on either corpus. The control ends below its start on both new corpora, at −0.030 [−0.063, +0.003] on HaluEval, where the interval reaches zero, and −0.193 [−0.247, −0.140] on GSM8K, where it is the worst arm, below the blind judge. The blind judge ends below its starting point on HaluEval, at −0.037 [−0.070, −0.007], and has the same point estimate on GSM8K, with an interval that reaches zero, [−0.077, 0.000]. On HaluEval the blind judge saw the question and the answer and lacked only the source knowledge, the configuration of a deployed judge that sees the user's prompt but not the material the system retrieved, so this is the realistic version of the harm, although it is not distinguishable from the control (−0.037 against −0.030). The informed Gemini judge does not rescue either corpus, although a second judge family does (below). On HaluEval it gains +0.013 [−0.013, +0.040], within noise, and on GSM8K it ends below its start at −0.020 [−0.067, +0.030], fixing 23 solutions and breaking 29, with a report precision of 0.199 against 0.139 for the blind judge. Seeing the problem statement did not make the judge's complaints about arithmetic reliable, and only the answer-key certificate improved GSM8K (+0.030 [+0.013, +0.050]). On GSM8K, an instruction to revise with no content broke 69 of 265 correct solutions. We explain this reversal by what a second pass can achieve on each corpus. An IFEval constraint can be re-checked by re-reading the instruction, so revision has something to converge on. A finished arithmetic solution has nothing further to consult, so an unmotivated instruction to revise invites the model to change an answer that was already correct. Where revision has no evidence to return to, revision pressure alone was the most destructive regulator we measured. The counts are measurements, whereas this account of the mechanism is an interpretation. On this account the damage has two components: revision pressure, scaled by the fragility of the task, and wrong reports, scaled by one minus precision.

Two exploratory runs, made after the committed rounds and not counted in Table 6, test the closed-loop result where it would be most likely to fail. In the first we hid IFEval's constraints from the judge, as if they sat in a system prompt. On the same 300 responses, the judge saw the response together with the user's request, rewritten once by Gemini 2.5 Flash-Lite with every requirement on format, length, case, keywords and structure removed. Acting on its reports changed the pass rate by −0.080 [−0.119, −0.041], breaking 66 constraints while fixing 28. That is 13.9 points below the no-report control [−0.174, −0.102] and only slightly less harmful than the judge shown the response alone (a difference of +0.040 [0.000, +0.079]). Two further runs, with fresh judge votes and revisions, gave −0.080 and −0.084. A judge that sees the user's request but not the requirements stated elsewhere does nearly the same damage.

In the second run we replaced the feedback judge with a second model family, Claude Haiku 4.5, and kept the responders' original outputs, the no-report control and the certificate. The blind judge again left the system below its start on all three corpora, at −0.092 [−0.130, −0.054] on IFEval, −0.073 [−0.117, −0.030] on HaluEval QA and −0.037 [−0.073, 0.000] on GSM8K. Unlike the informed Gemini judge, the informed Haiku judge improved all three, by +0.071 [+0.042, +0.102], +0.157 [+0.113, +0.203] and +0.057 [+0.020, +0.093]. Two further runs of every arm kept every sign: the blind judge ended at −0.080 and −0.120 on IFEval, −0.063 and −0.067 on HaluEval QA and −0.033 and −0.020 on GSM8K, and the informed judge at +0.065 and +0.050, +0.150 and +0.157, and +0.047 and +0.060. The harm of blind feedback replicates across judge families and across runs, whereas the benefit of informed feedback depends on the judge. On HaluEval QA and GSM8K the informed Haiku judge also gained more than the certificate, because on those corpora the certificate states only that the answer is wrong and names no defect. The HaluEval figures are scored as in Table 4, by containment and then entailment, and the same scoring reproduces Table 4's Gemini arms exactly, so the two judge families can be compared directly.

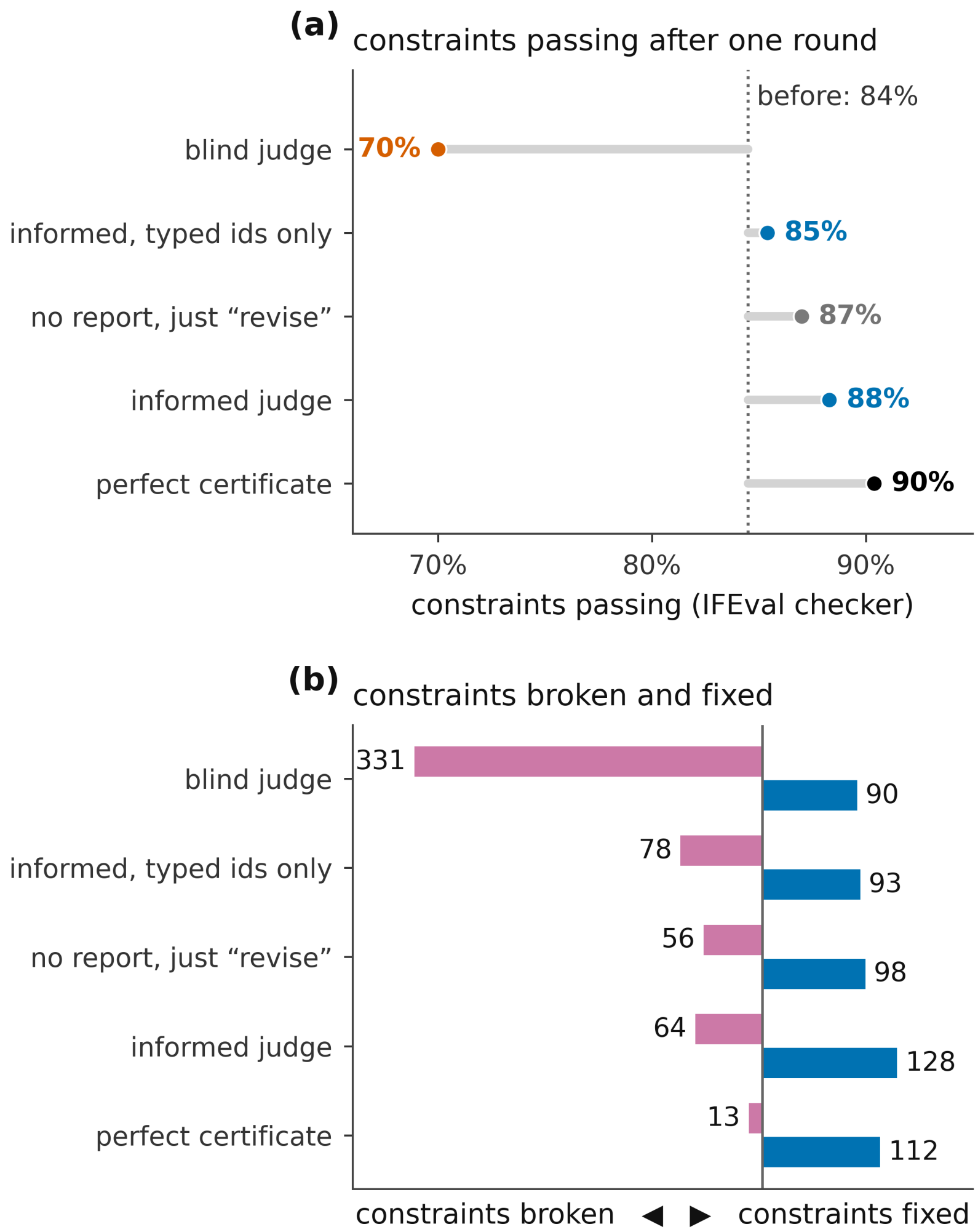

**Figure 15:** Acting on a blind judge's reports breaks more than it fixes. IFEval: 1,080 responses, 1,664 constraints, one repair round, checker-scored. (a) Constraints passing after the round (dotted: before). (b) Constraints fixed and broken per regulator, rows as in (a).

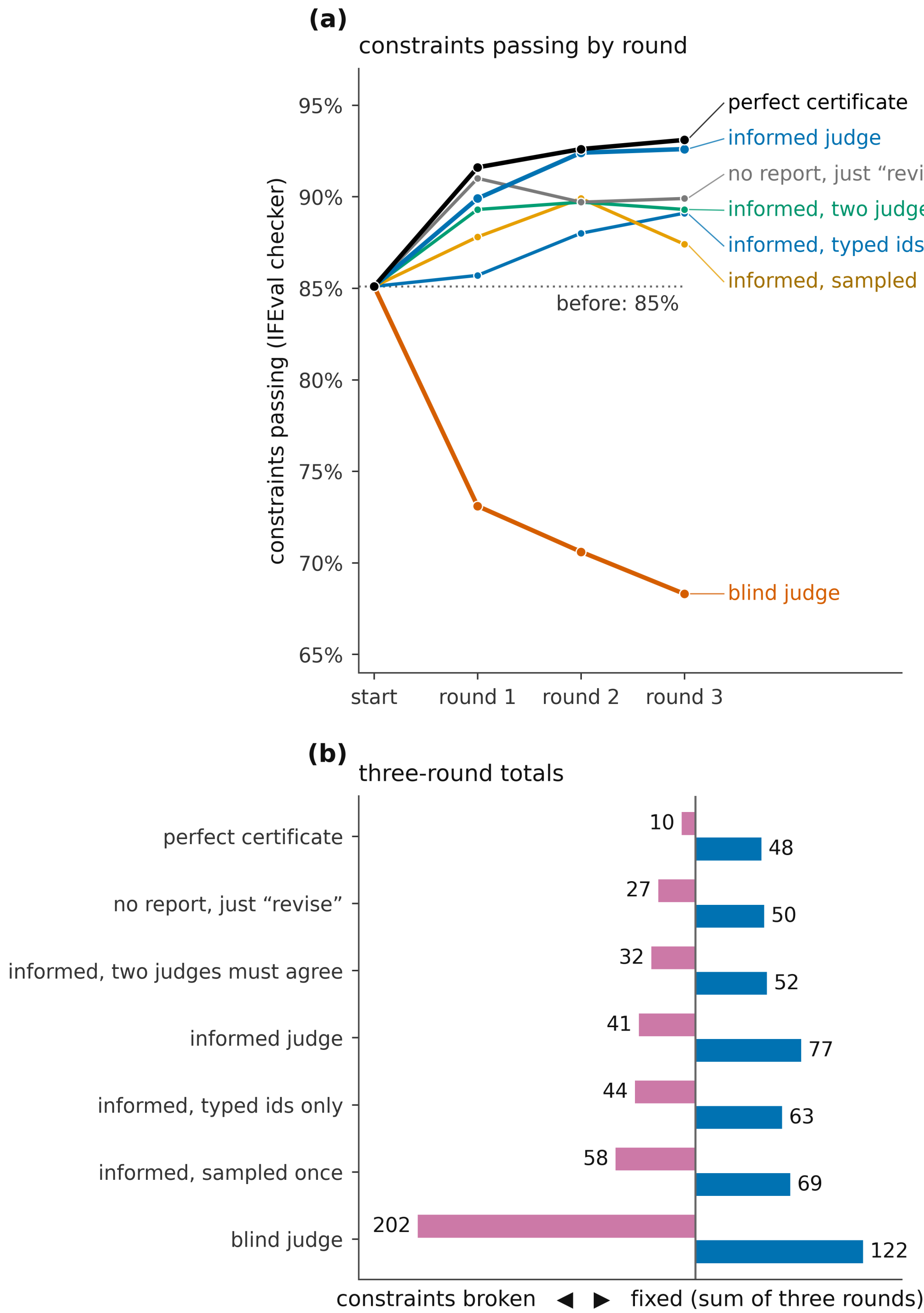

**Figure 16:** Each round of blind feedback leaves the system worse than the last. The 300-response subset iterated for three rounds with fresh judging each round. (a) Constraints passing per round. (b) Constraints fixed and broken, summed over the three rounds. "Informed, sampled once" samples the informed judge once instead of three times. "Informed, two judges must agree" acts only on defects reported by both the free-text and the typed-id informed judges.

**Table 4:** On all three datasets, a blind judge never rises above where it started and a perfect certificate lifts the system most. On GSM8K the informed judge also ends below its start. §8.2 reports the same loops with a second judge family. The same repair loop is run on each dataset, scored by its own programmatic checker rather than by a judge. Intervals are given in §8.1 and §8.2.

| after one repair round | IFEval (constraints passing) | HaluEval QA (answers correct) | GSM8K (solutions correct) |
|---|---|---|---|
| before the round | 84% | 77% | 88% |
| blind judge | 70% | 73% | 85% |
| informed judge, typed report | 85% | — | — |
| informed judge | 88% | 78% | 86% |
| no report, just "revise" | 87% | 74% | 69% |
| perfect certificate | 90% | 81% | 91% |

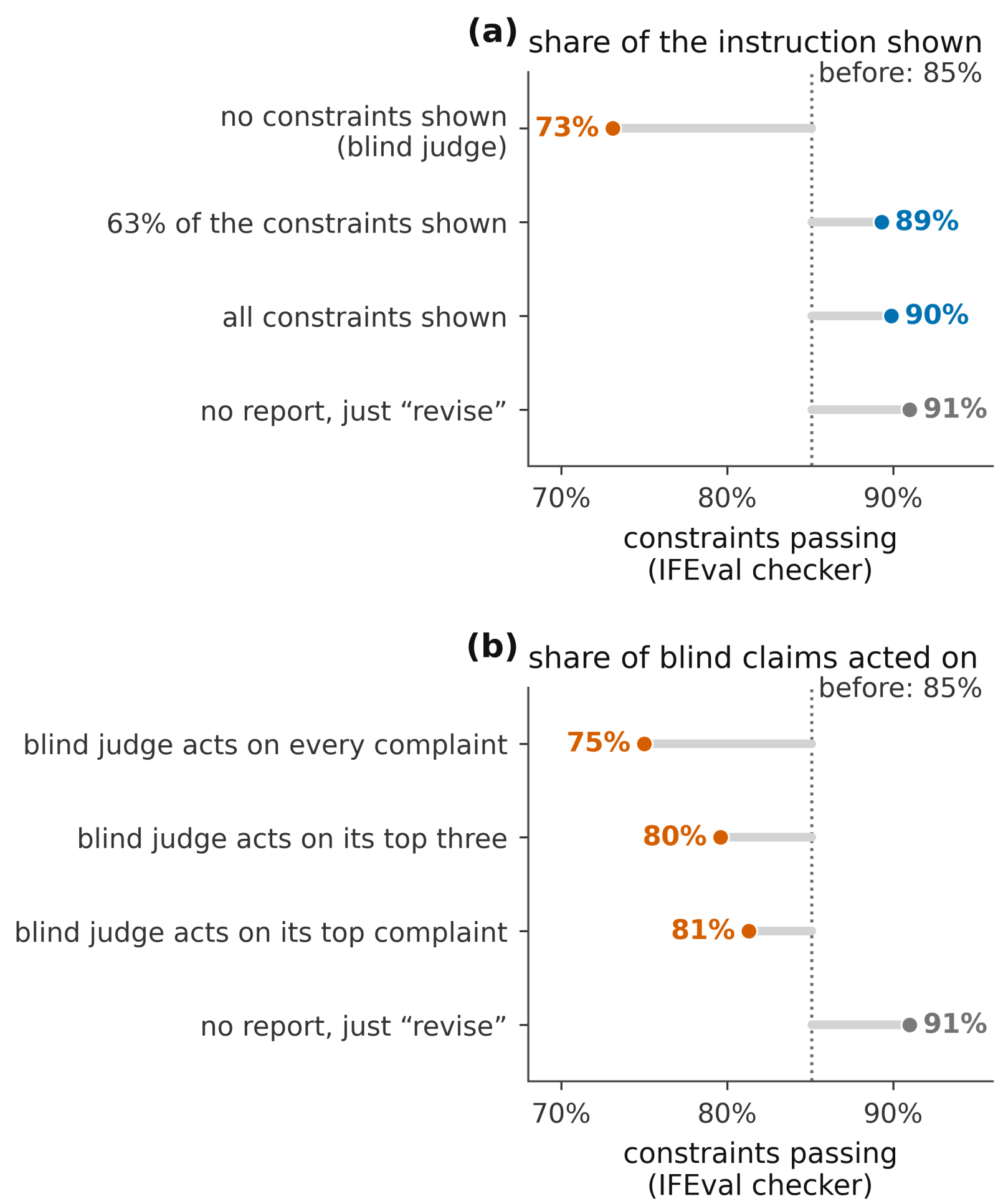

**Figure 17:** Most of the instruction's value arrives with 63% of it; throttling a blind judge does not remove the harm. IFEval 300-response subset, one repair round. (a) Share of the instruction shown to the judge: none, a random subset (63% of constraints as realised), all. (b) Share of a blind judge's claims acted on: the top one, the top three, all. Table 4 reports the same loop on three datasets.

## 9 Do the same rules govern the agent? (RQ6)

We test the same manipulations on the generating side, to see whether the world handed to the generator regulates it. We use HaluEval QA with 500 questions and two agents, Llama-3.1-8B (the weaker) and Claude Haiku 4.5 (the stronger), under the same prompt and temperature. Correctness is scored by containment of, or entailment to, the corpus answer, with no judge in the loop. We construct seven worlds, each of which varies one of Ashby's properties: no world (W0); another item's knowledge of the same length (W1, variety without a model); a random half of the item's knowledge (W2); all of it (W3); all of it with three distractor documents (W4, purity); all of it as numbered facts keyed to the question (W5, structure); and the plausible false answer presented as the knowledge (W6, a model of the wrong system).

Both agents show the same curve (Figure 18). With no world, accuracy is 0.316 for the weaker agent and 0.446 for the stronger, rising to 0.780 and 0.812 with half the knowledge and to 0.926 and 0.948 with all of it. Distractor documents and numbered facts change accuracy little, whereas irrelevant knowledge is worse than none for both agents, at 0.200 and 0.204. The stronger agent starts 13 points higher and still gains 50 points from the world, and the weaker agent gains 61. We had committed to the prediction that a plausible false world would fall below no world, which held for Haiku (0.328 against 0.446 with no world) but not for Llama (0.358 against 0.316). Our interpretation, which we have not measured, is that the 8B agent's own model is weak enough that a plausible false world does about as well as none.

Grading the world for each item is harder than grading it for each condition, which points to a requirement we state but do not meet here. A regulator that revises its own model needs to sense when that model is no longer sufficient, and so needs some measure of the gap between the world it holds and the world its subject is working in. Without such a measure it cannot know when to revise. Our first attempts failed to resolve individual items. A settlement margin motivated by the theory recovered the ordering of the conditions but predicted correctness at $\rho = 0.17$, under its committed bar of 0.25, and plain entailment adequacy reached a pooled $\rho$ of 0.28–0.31 (post hoc) but only 0.03–0.13 within the full-world conditions. We leave such a sensor to future work.

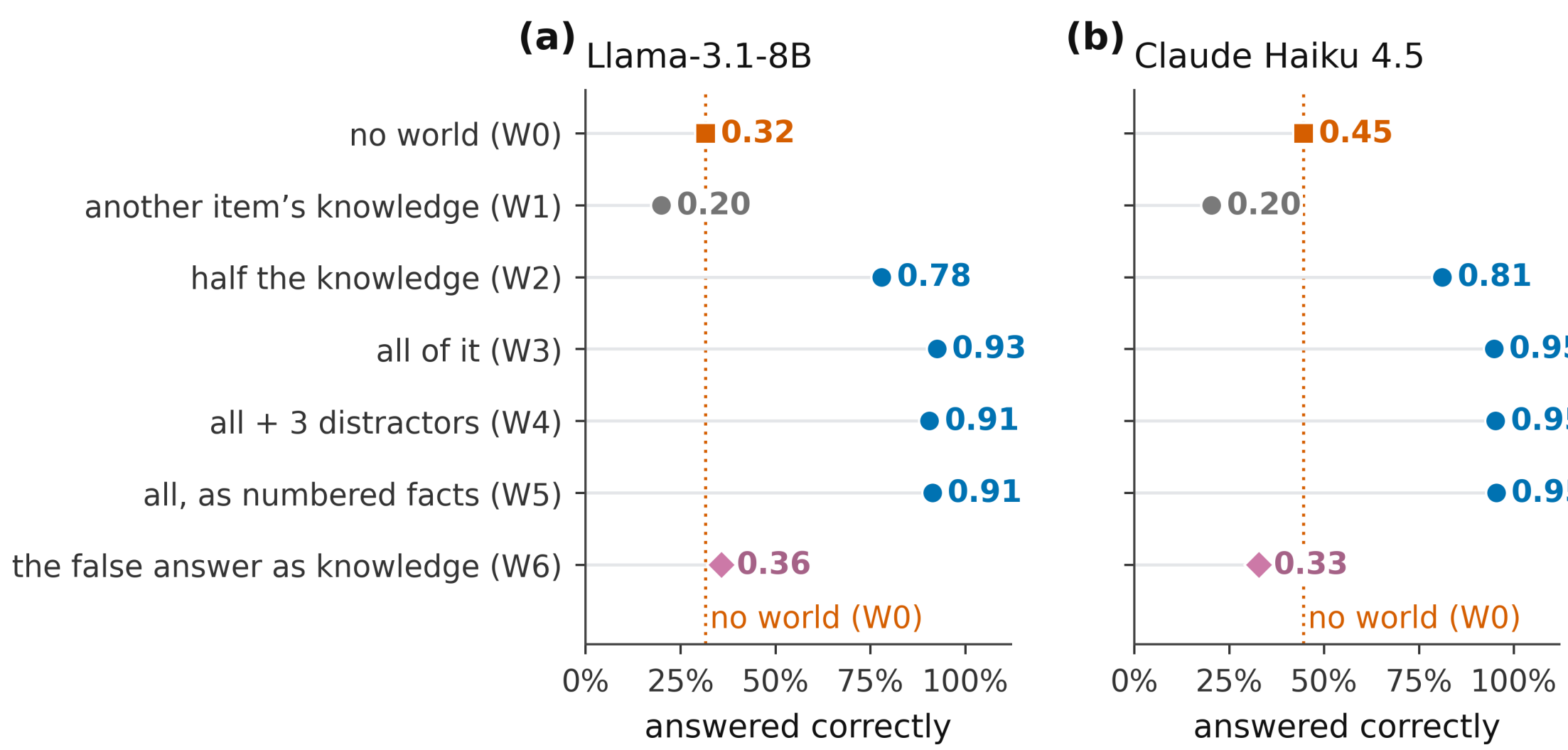

**Figure 18:** An agent answers only as well as the material it is handed. HaluEval QA, 500 questions, correctness scored against the corpus answer with no judge in the loop. (a) Llama-3.1-8B and (b) Claude Haiku 4.5 under the seven worlds of §9 (W0–W6); dotted line: accuracy with no world.

## 10 The design rules, and what it costs to get them wrong

### 10.1 The rules

Table 5 collects the design rules that follow from these results, and Figure 19 arranges them by the four design questions. Each rule is a measurable requirement on the judge as regulator and is paired with the measurement that supports it. Read together, the rules answer the four design questions of §1: which channels to open, how much of each to supply, in what form, and what the reports must satisfy. The quantities in the table are the effects we measured on these corpora, and the framework is the procedure for establishing them afresh on a new task. §11 records where a single run is all that supports a number.

**Table 5:** The design rules this paper establishes, each with the experiment that measures it. Each effect size is followed by the section that reports it, and every rule rests on at least one prediction committed before its run.

| Rule | Because | Measured |
|---|---|---|
| Give the evaluator the system's own evidence, or give it nothing | a judge given plausible but wrong material raises more false accusations without finding more errors | placebo source: false positives 12.4 → 20.9 per error-free answer; irrelevant world for an agent: accuracy 0.20 against 0.32 / 0.45 with none (§5, §9) |
| On open-ended tasks, do not supply a reference answer as evidence | on summarisation a reference shows one acceptable output among many and says nothing of the source, yet the judge relies on it over the source; on single-answer questions it helped | balanced accuracy 0.552 against 0.565 blind; beside the article it removes 82% of the precision gain from the article (FRANK, three runs); on HaluEval QA it reached 0.919 against 0.852 with the knowledge (exploratory, three runs; §5) |
| Route the evidence to the claims; do not pre-digest it | the sentences nearest each claim carry almost all of the value, while a restatement of the source loses some | relevant 11% keeps 98% of recall; a faithful restatement costs balanced accuracy 0.700 → 0.665 (§6) |
| Put the structure on the requirements, not the evidence | typed requirements improve precision; taxonomies and routed evidence do not | Shapley −3.59 false positives for typed requirements; taxonomy (IFEval) +7.7 recall at +19 points of false-alarm rate (§6) |
| Require a citation per defect | it gave the cheapest precision gain we measured | false positives 3.36 → 2.88 at a 0.5-point recall cost (§6) |
| Judge an evaluator by precision, not recall | a judge can raise recall by flagging everything, whereas precision determines whether acting on its reports helps | structural judge with no source: recall 1.000 at 9.45 false positives (§6) |
| Do not act on a bare score | a score names no defect, so the responder revises without knowing what to fix | scalar report +0.036 against +0.059 for the no-report control (§8) |

| Rule | Because | Measured |
|---|---|---|
| Fix what counts as a defect outside the evaluator | the checker's certificate was the most precise regulator and acted on the fewest responses, whereas making the judge act on fewer responses did not help | certificate acts on 236 of 1,080 responses and breaks 13; on the 300-response subset a stated goal cuts responses acted on from 275 to 228 with no net gain (§8) |
| Grade the world given to the agent, not only its output | the same agent scores 0.20, 0.45 or 0.95 depending on what it was handed | §9, both agents |

The precision rule has a degenerate form, and the last of the report rules guards against it. A judge that optimises only for finding defects can raise recall by flagging everything, and our data contain an instance in the structural judge with no source, which attains recall 1.000 at 9.45 false positives by calling every claim unsupported. A judge that optimises only for precision can reach it by reporting almost nothing, so that it is never wrong but also protects nothing. Our quietest arm, the certificate, avoids both because a checker outside it fixes what counts as a defect, which is why it acts on only 236 of 1,080 responses. Making the judge itself quieter did not have the same effect, since stating the goal reduced the responses it acted on from 275 to 228 without improving the net outcome (P37). The essential variables must be set outside the evaluator, because otherwise an evaluator could meet any precision target simply by reporting less.

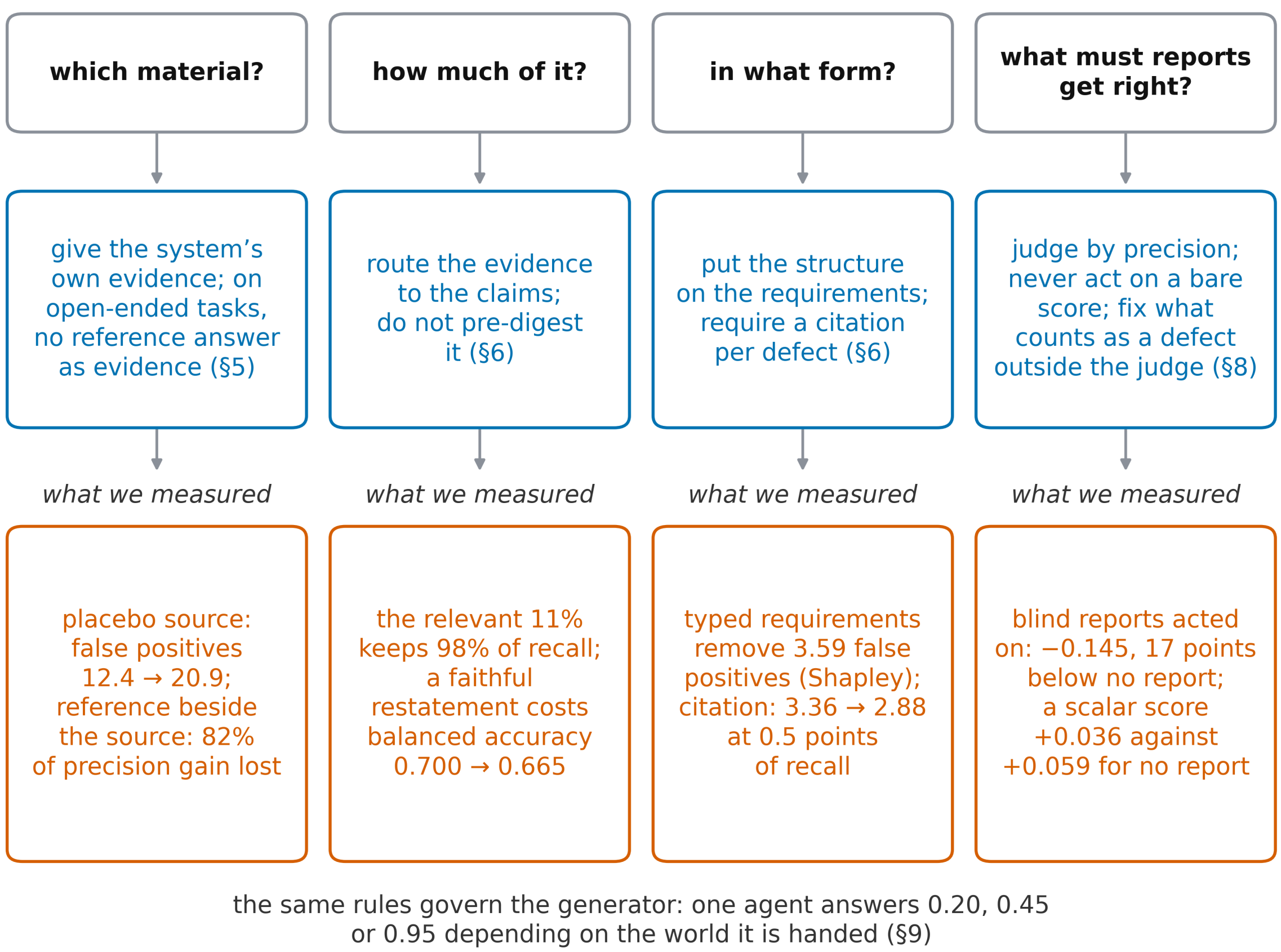

**Figure 19:** The design rules, and what the measurements showed. For each of the four design questions of §1, we show the rule each measurement supports (Table 5) and the effect measured on these corpora.

### 10.2 What the theory predicted, and what it did not

The framework posed the questions and generated most of the predictions, but it did not generate most of the quantitative results, and we set out here where it failed. Every prediction that turned an information-theoretic quantity into a per-item measurement failed or came out null. The four such clauses that carried a FAILS condition, on answer variety (P17(a)), on the answer-partition entropy $H(A|q)$ and the source information $I(A;S|q)$ (P19(b), P20), and on cardinality class (P22(a)), gave rank correlations between −0.10 and +0.02 where their bars asked for +0.15 to +0.30, in part because most questions sat at the ceiling of the entropy measure (Table 6). The one coded quantity that tracked the paired gain was the evidence-location code of §4 (P4), and it clears its bar only when the two reversed translation pairs are included (§11).

Two structural predictions also failed. The four-factor Shapley ordering of P14(a) failed, and the result reported in §6 is a three-factor decomposition within the language-model judge. The regulator ordering of P24(a) on IFEval failed, and the precision ordering of §8.2 was read from those data after the failure. It was then committed for HaluEval QA and GSM8K (P35(c)) and held, but each of those orderings has three arms, one of them the oracle, so a perfect rank correlation has $p = 1/6$ on each corpus. The predictions about the generating side were mixed. The dose-response of §9 held for both agents, the false-world prediction held for one of the two (P28(a)), and the per-item adequacy and settlement-margin sensors failed their bars (P28(b), P33(a)).

What held were mostly qualitative contrasts: the gap between blind and informed judges across corpora and judges (P1, P2, P5(a), P29–P31), the harm of acting on a blind judge's reports (P24(b), P25(a)), the failure of the reference arm on FRANK (P12, P15) and the variety sweep (P34(a)). The Ashby framing organises these findings and suggested where to look, but the findings stand on their measurements, and the framing should not be read as having produced them. Of the 81 committed clauses that ran, 32 met their bars and 50 went in the predicted direction. The exploratory runs of §5 and §8.2 and the matcher audit of §3.6 were made after the committed rounds and are not counted in these figures.

## 11 Limitations

The results that depend on the matcher, the small language model that decides whether a complaint points at a marked segment, carry the uncertainty described in §3.6. On RAGTruth the two matching routes disagree in sign on the paired contrast (+0.263 against −0.028, Figure 5b), and the routes agree less often on the blind arm (0.70) than on the informed arm (0.94), an asymmetry that runs in the direction of our effect. The mechanical audit of §3.6 bounds it only in part: on RAGTruth a second matcher family reproduced the stored decisions on 92% of the blind arm and 96% of the informed arm, and a permuted control found 1% false matches, but no person has read the decisions. Every claim that depends on the matcher has a counterpart that does not: the corpus-protocol and LLM-AggreFact ladders (§4), the label-scored corpora (§4, §5) and the checker-scored loops (§8). Those are the results to weigh. No scoring decision was audited by a human.

Some measurements contain model annotation. The question profile of §5 is a language model's reading of each question, and the typed-versus-free-text contrast routes free text through an automatic keyword table. HaluEval correctness in §8 and §9 falls back to an entailment model where string containment fails, so on HaluEval "automatic" does not mean "programmatic" as it does for IFEval and GSM8K. The restatement-fidelity statistic of §6 and the world-adequacy grade are also entailment outputs.

Our replication is limited in several ways. The committed closed loops used one feedback judge, Gemini 2.5 Flash. The second-family replication of §8.2 is exploratory; it was run three times per corpus and kept its signs each time. We used three commercial judges, two of them from the same tier, and stopped a fourth judge, Claude Sonnet 5, for cost (§3.8). The out-of-English reversal was measured with one judge family. The following are each a single run with one dataset, one judge family and one responder pair: the factorial, the evidence-delivery arms of §6, the restate-then-judge test, the question-demand profile of §5, the settlement margin, the sweeps, the three-round loop and the two new closed loops.

Bootstrap intervals are descriptive rather than significance tests, and where an interval crosses zero we say so. The 49-category rank test permutes categories as if they were independent, although they are nested in corpora, and it clears its committed bar of 0.6 only when the two reversed pairs are included. Stratum-level claims in §4 inspect 22 strata (11 on RAGTruth and the 11 LLM-AggreFact sub-datasets) without multiplicity correction. The question-demand effect is at $p = 0.052$.

Table 7 in Appendix B lists every run with its corpus, sample, judge, scoring and pre-commitment status. The ladder, placebo and dilution arms use the first 250 RAGTruth items. We compare LLM-AggreFact, a balanced 1,100-item sample, against published full-test-set figures, so the comparison is indicative rather than like-for-like. The three-round loop, the sweeps and the channel experiment use a 300-response subset, on which the no-report control gains more (+0.059) than on the full corpus (+0.025). All of these choices were recorded before the results.

The reference-answer results rest on one summarisation corpus and one question-answering corpus, each run three times with the same outcome, and the second is exploratory. The hidden-constraint arm of §8.2 is exploratory, was run three times on one corpus, and its constraint-free requests were written by a language model rather than by hand. Two of Ashby's requirements are only partly isolated. An explicit goal set improves precision but not net gain, as measured on one corpus and one round (§8). The requisite-variety requirement rests on a single sweep, in which the "half" arm in fact showed 63% of the constraints, in a different representation from the full arm.

## 12 Conclusion

We begin with the closed-loop result, because it matters most for deployment. A judge denied the material its subject had to respect is not, in a repair loop, an inexpensive approximation of an informed one, since it left the system below where it started on two of three corpora and within noise of it on the third. On instruction following it performed worse than providing no report, and the loss grew with each round. This matters because a wrong report that is acted on introduces new defects. That result reflects a more general one, namely that the limiting factor in LLM-based evaluation is whether the evaluator's channels carry the variety needed to separate the verdicts the task requires. An evaluator denied the world its subject was working in cannot see defects whose evidence lies there, and in our experiments additional judge capability did not recover the difference. Nor is the repair simply "more information". What helped was the system's own evidence, routed to the claims and left whole, together with requirements stated as types, defects that cite their evidence, and a precision standard for every report. We found that the same rules govern the agent that produced the output, so the world given to an agent should be evaluated along with what the agent did with it. Our claim is scoped to corpora in which the evidence of a defect lies outside the output. Where it does not, supplying the source can cost accuracy, as it did on the two out-of-English translation pairs (§4).

These results suggest a standard for evaluators, and we propose that an evaluator is sufficient only if three conditions hold. First, its model must track the variety that reaches the verdict. Second, it must revise that model when the system changes, both on announced changes and on errors already made. Third, it must measure its own model–system gap, so that it knows when revision is due.

We establish the first condition here and leave the other two, which are at different stages, to future work. The second is untested here. Every loop we ran revises on error, which is the easier half, but none revises on advance notice that the system it grades has changed. We have started on the third. The sufficiency measures of §9 recover the ordering of the worlds, but in their initial form they predict correctness within a condition at ρ 0.03–0.13. That signal resolves the condition but not yet the item, which is the immediate next target. Ashby uses the term ultrastability for a regulator that reorganises itself when its essential variables leave their bounds [26, 57]. The important property is that the trigger for that reorganisation lies inside the regulator. A quarterly review of an evaluation suite can be seen as an ultrastability loop whose trigger is placed outside it, and if such loops decay, that placement may be the reason. Building a loop with an internal trigger is the next requirement, and it depends on the per-item sufficiency sensor.

## Data and code availability

The seven corpora used here are public, and we use them as released, apart from the IFEval responses described in §3.2. No data collected by us is used as evidence anywhere in this paper, and every experiment can be reproduced from the corpora together with the protocol in §3. The prediction ledger with its commit hashes and per-round snapshots, the verification log, every judge prompt and per-run result artefacts are available from the authors on request.

## Ethics and broader impact

This paper studies automated evaluators, and its central negative result carries a deployment risk that we should state directly. A judge that cannot see the world it is judging misses defects whose evidence lies in that world and, when its reports are acted on automatically, also introduces defects (§8). As a result, systems that route low-precision evaluator feedback into automated revision can degrade below taking no action at all, and our multi-round results suggest that the harm grows with the degree of automation. We recommend that closed-loop deployments verify repairs against an instrument outside the judge–responder loop, as §3.5 does here. All corpora used are public research datasets employed for their intended purpose, and none contains personal data collected by us. The generating models and judges are commercial and open-weight systems accessed through their public interfaces. No human subjects were involved, and no human annotator labour was commissioned for this work. Every label we use is one the corpora already publish.

## Use of AI assistants

We used language models for code assistance and for editorial revision of the manuscript. We did not use them to generate experimental results, to select or interpret findings, or to write the analysis. Every number reported comes from the run artefacts, and the authors take responsibility for the entire content.

## Acknowledgements

We thank the authors and annotators of the seven public corpora used here, without whose released labels none of these measurements would be possible.

# Appendix A Every committed prediction

Prediction commits by round (short hash, date): 1: 9dad951, 2026-08-28; 2: cfc8009, 2026-08-29; 3: aa484d1, 2026-08-29; 4: ad59290, 2026-08-29; 5: 4bd23bc, 2026-08-29; 6: 22760d5, 2026-08-29; 7: bec03ed, 2026-08-29; 8: 9c64861, 2026-08-29; 9: a779f7e, 2026-08-29; 10: 4c2e618, 2026-08-30; 11: 5d79802, 2026-08-30; 12: e920452, 2026-08-30; 13: 809c93e, 2026-08-30; 14: 018a109, 2026-08-30; 15: 6f5f7f6, 2026-08-31; 16: 2dfb230, 2026-08-31; 17: 6e57cd8, 2026-08-31. Full hashes and the SHA-256 of the ledger at each commit are available from the authors on request. These commits were made in the authors' private repository and had not been pushed to any remote when this version was prepared; no third-party timestamp exists for them.

**Table 6:** Of the 83 units in the committed clauses of the 37 predictions, 32 were met, 28 were not met, 18 triggered their FAILS clause, 3 reversed and 2 were not run. Each clause was written to a file and committed by hash before its runs, and its verdict is given against the threshold as written. The bar is quoted in short form; the full wording of each clause and of its FAILS condition is in the prediction ledger, available from the authors on request. Verdicts: met; not met (the FAILS clause was not triggered); FAILS (the FAILS clause was triggered); reversed (sign reversed with an interval excluding zero); not run. Where the observed direction differs from the verdict's default it follows the semicolon. FA denotes false alarms, the false positives of the body text; in the committed clauses, source-fed means informed and clean means error-free. No threshold was revised after seeing a result.

| # | Committed clause (bar as written) | Verdict; direction | Observed |
|---|---|---|---|
| P1 | blind recall rises with tier; strongest blind ≥ 10 pts below mid-tier source-fed 0.913 | met | blind 0.610 / 0.614 / 0.788; source-fed 0.924 / 0.906 / 0.969 |
| P1 (Sonnet 5 rung) | fourth rung of the ladder | not run | stopped at 30/400 items, projected ~$50 over budget |
| P2 | paired gain E−B ≥ +0.15, McNemar $p < 10^{-6}$, positive in every task and generator | met | +0.227 [0.177, 0.274], $p = 10^{-22}$, 6/6 |
| P3 | omission gain > mistranslation > fluency in both en-de and en-ru | reversed; opposite | source-fed below blind: en-de −0.074 ($p = 6\times10^{-5}$), en-ru −0.107 ($p = 4\times10^{-13}$) |
| P4 | Spearman ρ ≥ 0.6, permutation $p < 0.01$, categories n ≥ 8 | met | ρ 0.646, $p < 10^{-4}$, n = 49 |
| P5(a) | blind bal-acc ≤ 0.60; evidence arm ≥ 0.72; gain positive in ≥ 9/11 | met | 0.575 → 0.795, +0.221 [0.185, 0.257], 11/11 |
| P5(b) | structural judge within 0.05 of MiniCheck-FT5 0.750 | FAILS; opposite | 0.621 [0.596, 0.647], trails by 0.129 |
| P6 | blind violation recall ≤ 0.35; instruction-fed ≥ 0.85 | not met | blind 0.375; instruction-fed 0.779; gain +0.40 |
| P6 (structural clause) | structural requirement patterns recall ≥ 0.60 on propositional constraint types | not run | not run in this batch |
| P7 | blind bal-acc ≤ 0.65 and knowledge-fed ≥ 0.80 in each of QA, dialogue, summarisation | FAILS; null | QA +0.207 [+0.176, +0.239]; dialogue +0.025 [−0.005, +0.052]; summarisation +0.011 [−0.028, +0.047] |
| P8 | article-fed false-alarm rate lower than blind by ≥ 20 pts (XSum clean stratum) | not met | 12.5 pts, CI [0.02, 0.21] |
| P9 | union recall within [0.78, 0.91]; grammar the lowest article-independent category | met | 0.873 [0.805, 0.933]; GramE 0.818 < 0.932 |
| P10(a) | placebo recall within ±3 pts of B; false alarms higher than B | FAILS; opposite | recall 0.607 → 0.761 (+0.15 [+0.09, +0.22]); FA 12.4 → 20.9 |
| P10(b) | recall decays monotonically with 2/5/10 distractors; at 10 the gain is at most half | not met | 0.923 → 0.903 / 0.887 / 0.873; 85% of gain retained |

| # | Committed clause (bar as written) | Verdict; direction | Observed |
|---|---|---|---|
| P10(c) | structural judge under 5 distractors loses less than half the LLM's drop | met | −1.3 pts vs LLM −5.1 |
| P11 | typed arm recall ≥ untyped + 10 pts at equal or lower FA | not met; opposite on false alarms | +7.7 recall at +19 pts FA |
| P12 | reference-arm FA ≥ 15 pts above article arm; bal-acc ≥ 0.08 lower | met | FA +54 pts; bal-acc 0.552 vs 0.769 |
| P13(a) | routed top-5: bal-acc ≥ whole-document 0.700 + 0.05 | FAILS; null | 0.692 → 0.680, CI [−0.07, +0.04] |
| P13(b) | structural top-k = 5 improves bal-acc by ≥ 0.05 | not met | 0.522 → 0.560 |
| P14(a) | four-factor Shapley on recall: S largest, then T, then Q; $\lvert\varphi_I\rvert \le 0.05$ | FAILS; opposite in the four-factor values; as predicted within the language model | four-factor I +0.181, S +0.105 (I inflated by no-source accusation cells); LLM-only S +0.322 |
| P14(b) | T's recall gain with S present ≥ 2× its gain without S | FAILS; as predicted with S present | +0.015 with S, −0.107 without; the ratio is negative, so the test as written cannot be met |
| P14(c) | φ_I and φ_S negative on false alarms | not met; opposite on φ_I | φ_S −3.66; φ_I +1.54 |
| P15 | article+reference bal-acc within 0.03 of article-only or lower | met | 0.769 → 0.618; FA 0.305 → 0.695 |
| P16 | blind+taxonomy recall ≥ blind+free-text + 0.10 at higher FA | not met | +0.049 at +0.11 FA |
| P17(a) | per-item gain vs answer variety: Spearman ≥ 0.15, perm $p < 0.01$ | FAILS; opposite | ρ = −0.096 ($p = 0.97$) |
| P17(b) | blind false alarms vs answer variety: Spearman ≥ 0.15, $p < 0.05$ | not met; null | ρ = +0.09, $p = 0.19$ |
| P17(c) | QA gain ordered by answer-type class, open − polar ≥ 0.10 | not met; null (no variance) | 120 of 133 questions "open"; other classes n ≤ 4; no test possible |
| P18 | IFEval gain rises with constraint count: gain(3) − gain(1) ≥ 0.10 | FAILS; opposite | 1 → +0.57, 2 → +0.33, 3 → +0.34 |
| P19(a) | blind recall falls with H(A\|q): Spearman ≤ −0.15, $p < 0.05$ | not met; null | ρ = +0.06 ($p = 0.25$) |
| P19(b) | gain rises with I(A;S\|q): Spearman ≥ +0.20, $p < 0.01$ | FAILS; null | ρ = +0.02 ($p = 0.39$); 101/166 questions at entropy ceiling |
| P19(c) | blind false alarms rise with H(A\|q): Spearman ≥ +0.15 | not met | ρ = +0.24 ($p = 0.09$, n = 33): ρ clears the bar as written, but with $p = 0.09$ and the entropy measure at ceiling we score it NOT MET; the log's FAILED refers to the P19 bundle, whose (b) clause failed |
| P20 | HaluEval QA per-item gain vs I(A;S\|q): Spearman ≥ +0.15, $p < 0.01$ | FAILS; null | ρ = +0.02 ($p = 0.33$) |
| P21(a) | W2 recall within 0.03 of W1 and FA ≤ W1 | not met; as predicted on recall; opposite on FA | recall 0.908 (within 0.03); FA 3.73 vs 3.36 |
| P21(b) | W3 recall ≤ W1 − 0.15 | not met | 0.870 (W1 0.927): fell, by less than predicted |
| P21(c) | W4 FA ≤ 0.75 × W1 at recall within 0.03 | not met | FA 2.88 = 0.86 × W1; recall 0.922 |
| P21(d) | W5 recall ≥ W1 and FA ≤ 0.5 × W1 | not met; opposite on FA | recall 0.979 (met); FA 18.39 = 5.5 × W1 |

| # | Committed clause (bar as written) | Verdict; direction | Observed |
|---|---|---|---|
| P22(a) | cardinality class vs sampled H(A\|q): Spearman ≥ 0.30 on each corpus | FAILS; null (no variance) | ρ = −0.02 (RAGTruth), 0.02 (HaluEval); no variance |
| P22(b) | blind false alarms rise with cardinality class: Spearman ≥ 0.15 | not met; null | ρ = 0.04 (n = 33) |
| P22(c) | world gain larger for mention-all than mention-some by ≥ 0.08 | met | +0.389 vs +0.251, diff +0.138; bootstrap CI [−0.02, +0.30], perm p = 0.052 |
| P22(d) | explanatory questions highest blind FA per item; polar/individual lowest | met | explanatory 14.8 > individual 12.0 > polar 0 (n = 33 clean) |
| P23 | removing the question lowers blind bal-acc by ≥ 0.10 | reversed; opposite | 0.645 → 0.790 (+0.145) |
| P24(a) | net ordered R0 < R1 < R2 < R3 ≤ R4 | FAILS; opposite (R3 below R2) | R1 < R3 < R0 < R2 < R4; R3 +0.009 < R2 +0.038 |
| P24(b) | R1 broken ≥ R1 fixed and R1 net ≤ R0 net | met | R1 −0.145 [−0.168, −0.123] vs R0 +0.025; 331 broken / 90 fixed |
| P24(c) | R3 broken ≤ 0.7 × R2 broken | FAILS; opposite | 78 vs 64 |
| P24(d) | break rate on flagged-but-followed ≥ 2× unflagged, for R1–R3 | not met; as predicted for R2 only | R2 2.8×; R1 1.3×; R3 1.2× |
| P25(a) | R1 pass rate falls every round | met | 0.851 > 0.731 > 0.706 > 0.683 |
| P25(b) | R2 rises each round with shrinking increments, never falling | met | +0.048, +0.025, +0.002 |
| P25(c) | R4 round-2 and round-3 gains each < 0.01 | not met | 0.010, 0.005 |
| P26 | G net ≥ R2 net and G broken ≤ 0.7 × R2 broken | not met; null | +0.042 vs +0.048; 12 broken vs 11.2 bar |
| P27 | V1 net ≤ R3 net − 0.01 at round 1 | FAILS; opposite | V1 +0.027 vs R3 +0.006 |
| P28(a) | per agent W6 < W0 ~ W1 < W2 < W3; W4 ≤ W3; W5 ≥ W3 − 0.02 | FAILS; opposite for one of two agents | Llama W6 0.358 ≥ W0 0.316; Haiku W6 0.328 < W0 0.446; dose-response held for both |
| P28(b) | per-item correctness vs world adequacy: Spearman ≥ 0.30, p < 0.001 | FAILS; null | ρ −0.003 / −0.014 |
| P28(c) | Haiku W0 ≥ Llama W0 + 0.10; Haiku W3 − W0 ≥ 0.15; Haiku W6 < W0 | met | 0.446 vs 0.316; +0.502; 0.328 < 0.446 |
| P29(a) | every blind tier ≤ 0.65; every document-fed tier ≥ 0.75 | met | blind max 0.587; document-fed min 0.775 |
| P29(b) | best blind trails worst document-fed by ≥ 0.08 | met | 0.587 vs 0.775, gap 0.187 |
| P29(c) | blind bal-acc non-decreasing in tier within 0.02 | met | 0.575 → 0.565 → 0.587 |
| P30(a) | every judge: source-fed F1 > blind F1 and source-fed precision > blind | met | F1 0.193/0.185/0.217 vs 0.336/0.379/0.476; precision higher in all three |
| P30(b) | Opus blind F1 < Gemini source-fed F1 | met | 0.217 vs 0.336; gap +0.119 [+0.092, +0.153] |
| P30(c) | blind recall under the protocol < 0.85 for every tier | met | 0.736 / 0.687 / 0.746 |
| P31(a) | BQ F1 ≤ B F1 + 0.03 and BQ precision ≤ B + 0.02, every judge | met | dF1 +0.014 / +0.019 / −0.015; dP +0.006 / +0.008 / −0.014 |
| P31(b) | Opus BQ F1 < Gemini informed F1 | met | 0.343 vs 0.553 under the second-phase scoring of P30 (§3.8); gap [+0.167, +0.258] |
| P31(c) | full corpus: BQ recall ≤ B + 0.05; BQ FA ≥ B FA | met | 0.615 vs 0.641; FA 15.00 vs 14.03 |
| P32(a) | summarisation: R ≥ E + 0.05 bal-acc | reversed; opposite | R − E = −0.035 [−0.057, −0.013] |

| # | Committed clause (bar as written) | Verdict; direction | Observed |
|---|---|---|---|
| P32(b) | QA: $\|R - E\| \le 0.02$ | met | +0.006 [−0.013, +0.026] |
| P32(c) | summarisation: $R \ge C + 0.03$ | not met | $R - C = +0.020$ [−0.002, +0.042] |
| P32(d) | faithful restatements exceed unfaithful by ≥ 0.05 bal-acc | met | pooled +0.126 (within summarisation +0.024; genre-confounded) |
| P32(e) | BR ≤ blind + 0.02 and BR FA not lower than blind | not met; opposite in form (more conservative, worse) | BR 0.612 vs B 0.645 (−0.033 [−0.059, −0.006]); FA 0.430 < 0.538 |
| P33(a) | pooled W2–W4 Spearman(margin, correctness) ≥ 0.25, $p < 0.001$, each agent | not met | 0.165 / 0.177, $p < 10^{-4}$ |
| P33(b) | mean margin orders W6 < W1 < W2 < W3; W4 within 0.05 of W3 | not met | ordering recovered; $\|W4 - W3\| = 0.126$ |
| P33(c) | within W3, failures' mean margin below successes' by ≥ 0.10, each agent | not met | Llama +0.106; Haiku +0.053 |
| P33(d) | contrast control NLI(W ⇒ r) alone gives $\rho < 0.10$ | not met; opposite | $\rho$ +0.284 / +0.314 |
| P34(a) | net(V_half) ≥ net(V0) + 0.05 and net(V_full) ≥ net(V_half) | met | −0.120 → +0.042 → +0.048 |
| P34(b) | net(G1) ≥ net(G3) ≥ net(Gall) and net(G1) ≥ net(Gall) + 0.06 | met | −0.038 ≥ −0.055 ≥ −0.101; margin 0.063 [+0.029, +0.099] |
| P34(c) | net(G1) ≥ control − 0.05 while net(Gall) ≤ control − 0.12 | not met | G1 −0.038 vs bar +0.009 (own CI excludes 0); Gall −0.101 (met) |
| P35(a) | HaluEval: blind net ≤ 0 and ≤ control − 0.05; informed ≥ +0.05; oracle ≥ informed | not met | blind −0.037 [−0.070, −0.007]; control −0.030; informed +0.013; oracle +0.047 |
| P35(b) | net_blind(GSM8K) ≥ net_blind(HaluEval) + 0.05; informed − blind gap smaller on GSM8K | not met; null | −0.037 vs −0.037; gap 0.017 vs 0.050 |
| P35(c) | regulators order by report precision in both datasets | met | Spearman +1.0 in both (three arms each, one the oracle) |
| P36(a) | net(C0) < net(C1) and net(C0) < net(C2) | FAILS; opposite | C0 +0.036; C1 +0.059; C2 +0.013 |
| P36(b) | net(C0) ≤ control + 0.01 | met | C0 − control = −0.023 [−0.047, −0.002] |
| P36(c) | C3 > max(C1, C2) > C0 | met | +0.074 > +0.059 > +0.036 (C3 − C1 CI crosses 0) |
| P37(a) | net(E1) ≥ net(E0) + 0.03 | FAILS; null | −0.011 [−0.032, +0.011] |
| P37(b) | E1 FA rate on followed constraints lower than E0 by ≥ 5 pts | met | −0.148 [−0.216, −0.082] |
| P37(c) | E1 acts on fewer responses than E0 | met | 228 vs 275 |

# Appendix B  Every run

**Table 7:** Every run reported in this paper, with the section that reports it, the corpus and sample, the judge or instrument, how it was scored, whether the score depends on our matcher, and whether a prediction was committed before it. An arrow marks a closed loop, from the feedback judge to the responder.

| § | Run | Sample | Judge or instrument | Scored by | Matcher | Committed |
|---|---|---|---|---|---|---|
| §3.6 | Matcher audit (exploratory) | FRANK and RAGTruth: 300 blind + 300 informed decisions and 300 permuted pairs each | Claude Haiku 4.5 as second matcher; permuted control | agreement with the stored verdicts; false-match rate | is the matcher | no (after the committed rounds) |
| §4 | RAGTruth capability ladder, corpus protocol | 250 answers with errors (384 segments) + 50 error-free | Gemini 2.5 Flash, Haiku 4.5, Opus 4.8 | corpus character overlap | no | yes (P30, P31) |
| §4 | RAGTruth capability ladder, recall | same 300 answers | same three judges | generous matcher | yes | yes (P1) |
| §4, §7 | LLM-AggreFact ladder | 1,100 claims, 11 sub-datasets | same three judges; entailment graph | corpus labels | no | yes (P5, P29) |
| §4 | RAGTruth paired gain | 400 answers (562 segments) + 100 error-free | Gemini 2.5 Flash; Haiku 4.5 replication | generous and strict matchers | yes | Gemini run no; Haiku yes (P2) |
| §4 | Strengthened blind control | same 400 answers | Gemini 2.5 Flash + contradiction detector | generous matcher | yes | no |
| §4 | WMT-MQM | 3 language pairs; en-de and en-ru 500 error + 100 error-free segments each | Gemini 2.5 Flash | matcher against error marks | yes | yes (P3, P4) |
| §4, §5 | FRANK, four world states | 310 summaries, 632 sentences; reference arms three runs | Gemini 2.5 Flash | sentence labels via generous matcher | yes | yes (P12, P15) |
| §4, §5 | HaluEval, three genres | 500 pairs per genre | Gemini 2.5 Flash | corpus labels | no | yes (P7, P23) |
| §4, §6 | IFEval paired | 1,080 responses, 1,664 constraints | Gemini 2.5 Flash | released checker | no | yes (P6, P11, P16, P18) |
| §5 | RAGTruth placebo and dilution | 250 + 50 answers | Gemini 2.5 Flash; entailment graph | generous matcher | yes | yes (P10) |
| §5 | HaluEval QA reference arms (exploratory) | 500 pairs, three runs | Gemini 2.5 Flash | corpus labels | no | no (after the committed rounds) |
| §6 | RAGTruth QA factorial and deliveries | 166 answers (133 + 33 error-free) | Gemini 2.5 Flash; entailment graph | generous matcher | yes | yes (P14, P21, P22) |
| §6 | Restate-then-judge | HaluEval QA and summarisation, 500 pairs each | Gemini 2.5 Flash | corpus labels | no | yes (P32) |
| §7 | Structural judge | RAGTruth 390 answers; HaluEval summarisation 200 pairs; FRANK | entailment graph | matcher (RAGTruth), labels elsewhere | partly | yes (P5(b), P9, P13) |

| § | Run | Sample | Judge or instrument | Scored by | Matcher | Committed |
|---|---|---|---|---|---|---|
| §8 | IFEval loop, one round | 1,080 responses, 1,664 constraints | Gemini 2.5 Flash → Llama-3.1-8B / Qwen3-14B | released checker | no | yes (P24) |
| §8 | IFEval loop: rounds, sweeps, channel, goal | 300-response subset | same | released checker | no | yes (P25–P27, P34, P36, P37) |
| §8 | HaluEval QA loop | 300 questions | Gemini 2.5 Flash → Llama-3.1-8B | containment or entailment to the answer | no | yes (P35) |
| §8 | GSM8K loop | 300 problems | Gemini 2.5 Flash → Llama-3.1-8B | answer key | no | yes (P35) |
| §8 | IFEval loop, constraints hidden (exploratory) | 300-response subset, three runs | Gemini 2.5 Flash → Llama-3.1-8B / Qwen3-14B | released checker | no | no |
| §8 | Closed loops, second judge family (exploratory) | 300 per corpus: IFEval, HaluEval QA, GSM8K; three runs | Claude Haiku 4.5 → responders as above | checker, answer key, containment or entailment to the answer | no | no |
| §9 | Agent under seven worlds | HaluEval QA, 500 questions | Llama-3.1-8B, Haiku 4.5 as agents | containment or entailment to the answer | no | yes (P28, P33) |